\pdfoutput=1

\documentclass[11pt]{article}

\usepackage[]{EMNLP2023}

\usepackage{times}
\usepackage{latexsym}
\usepackage{amssymb}
\usepackage{makecell}
\usepackage{booktabs}
\usepackage{bbm}
\usepackage{graphicx}
\usepackage{multirow}
\usepackage[T1]{fontenc}
\newcommand{\cf}[0]{\textsc{CounterFact}}
\usepackage[utf8]{inputenc}

\usepackage{microtype}
\usepackage{amsmath}
\usepackage{inconsolata}

\title{Editing Large Language Models: Problems, Methods, and Opportunities}

\author{
Yunzhi Yao{$^{\clubsuit\spadesuit}\footnotemark[1]$}, Peng Wang{$^{\clubsuit\spadesuit}\thanks{~~Equal contribution.}$}, Bozhong Tian{$^{\clubsuit\spadesuit}$}, {\bf Siyuan Cheng}{$^{\clubsuit\spadesuit}$}, {\bf Zhoubo Li}{$^{\clubsuit\spadesuit}$}, \\{\bf Shumin Deng}$^{\heartsuit}$, {\bf Huajun Chen}$^{\clubsuit\spadesuit\diamondsuit}$, \textbf{Ningyu Zhang}$^{\clubsuit\spadesuit}\thanks{~~Corresponding author.}$,\\
 $^\clubsuit$ Zhejiang University 
$^\spadesuit$ Zhejiang University - Ant Group Joint Laboratory of Knowledge Graph\\
$^\diamondsuit$Donghai Laboratory
$^\heartsuit$ National University of Singapore, NUS-NCS Joint Lab, Singapore\\
  \texttt{\{yyztodd,peng2001,tbozhong,sycheng,zhoubo.li\}@zju.edu.cn}\\
  \texttt{\{huajunsir,zhangningyu\}@zju.edu.cn,shumin@nus.edu.sg}
  }

\begin{document}
\maketitle
\begin{abstract}
Despite the ability to train capable LLMs, the methodology for maintaining their relevancy and rectifying errors remains elusive. To this end, the past few years have witnessed a surge in techniques for editing LLMs, the objective of which is to \textbf{efficiently} alter the behavior of LLMs within a specific domain without negatively impacting performance across other inputs. This paper embarks on a deep exploration of the problems, methods, and opportunities related to model editing for LLMs. In particular, we provide an exhaustive overview of the task definition and challenges associated with model editing, along with an in-depth empirical analysis of the most progressive methods currently at our disposal. We also build a new benchmark dataset to facilitate a more robust evaluation and pinpoint enduring issues intrinsic to existing techniques. Our objective is to provide valuable insights into the effectiveness and feasibility of each editing technique, thereby assisting the community in making informed decisions on the selection of the most appropriate method for a specific task or context\footnote{Code and datasets are available at \url{https://github.com/zjunlp/EasyEdit}.}.
\end{abstract}

\section{Introduction}

\begin{figure}
    \centering
    \includegraphics[width=0.5\textwidth]{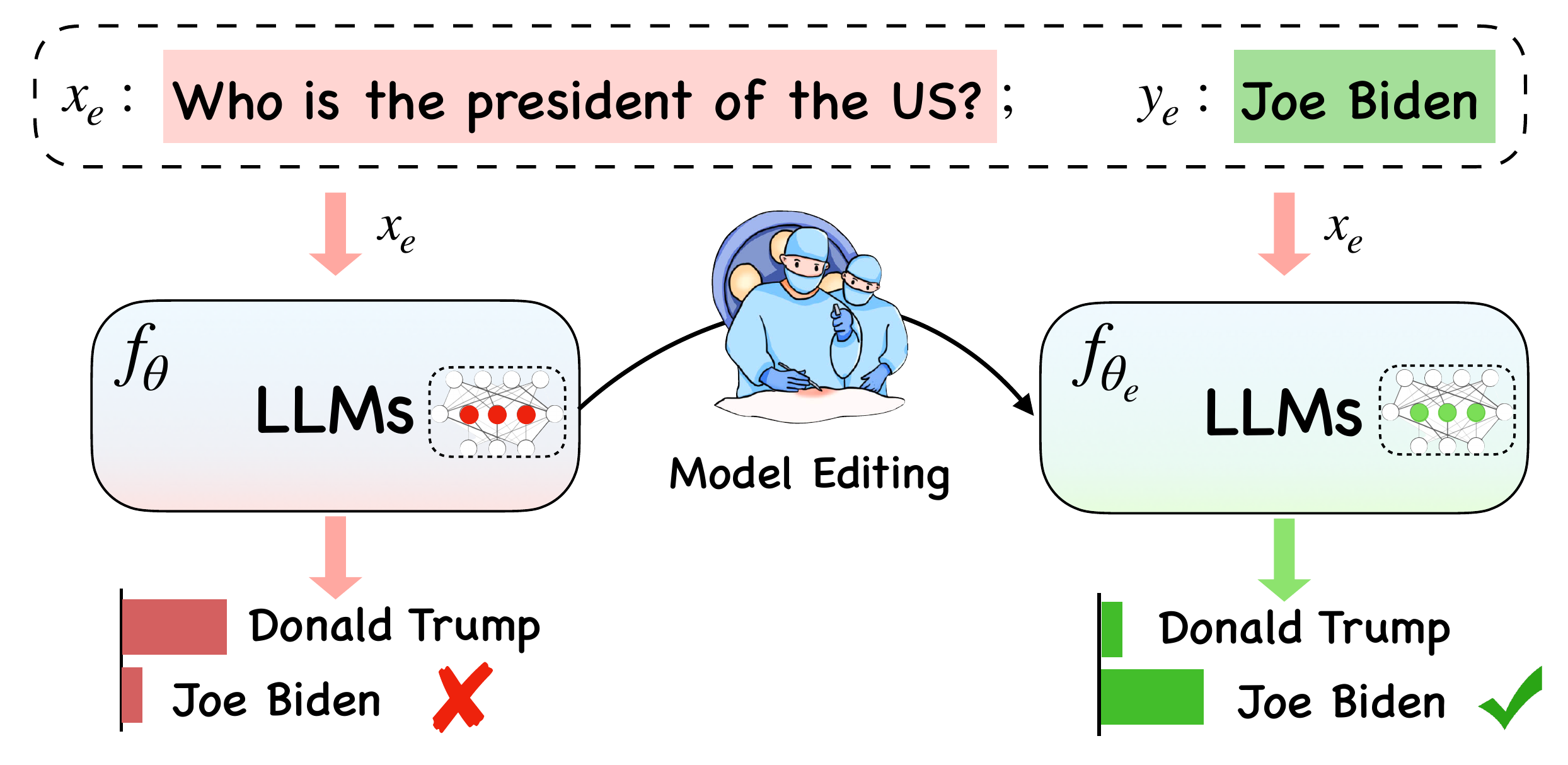}
    \vspace{-3mm}
    \caption{Model editing to fix and update LLMs.}
    \label{fig:intro}
    \vspace{-3mm}
\end{figure}

Large language models (LLMs) have demonstrated a remarkable capacity for understanding and generating human-like text \cite{DBLP:conf/nips/BrownMRSKDNSSAA20,DBLP:journals/corr/abs-2303-08774,anil2023palm,DBLP:journals/corr/abs-2302-13971,DBLP:journals/corr/abs-2212-09597,DBLP:journals/corr/abs-2303-18223}. 
Despite the proficiency in training LLMs, the strategies for ensuring their relevance and fixing their bugs remain unclear.
Ideally, as the world's state evolves, we aim to update LLMs in a way that sidesteps the computational burden associated with training a wholly new model. 
As shown in Figure \ref{fig:intro}, to address this issue, the concept of \textbf{model editing} has been proposed \cite{Sinitsin2020Editable,de-cao-etal-2021-editing}, enabling data-efficient alterations to the behavior of models, specifically within a designated realm of interest, while ensuring no adverse impact on other inputs.

\begin{figure*}
    \centering
    \includegraphics[width=0.92\textwidth]{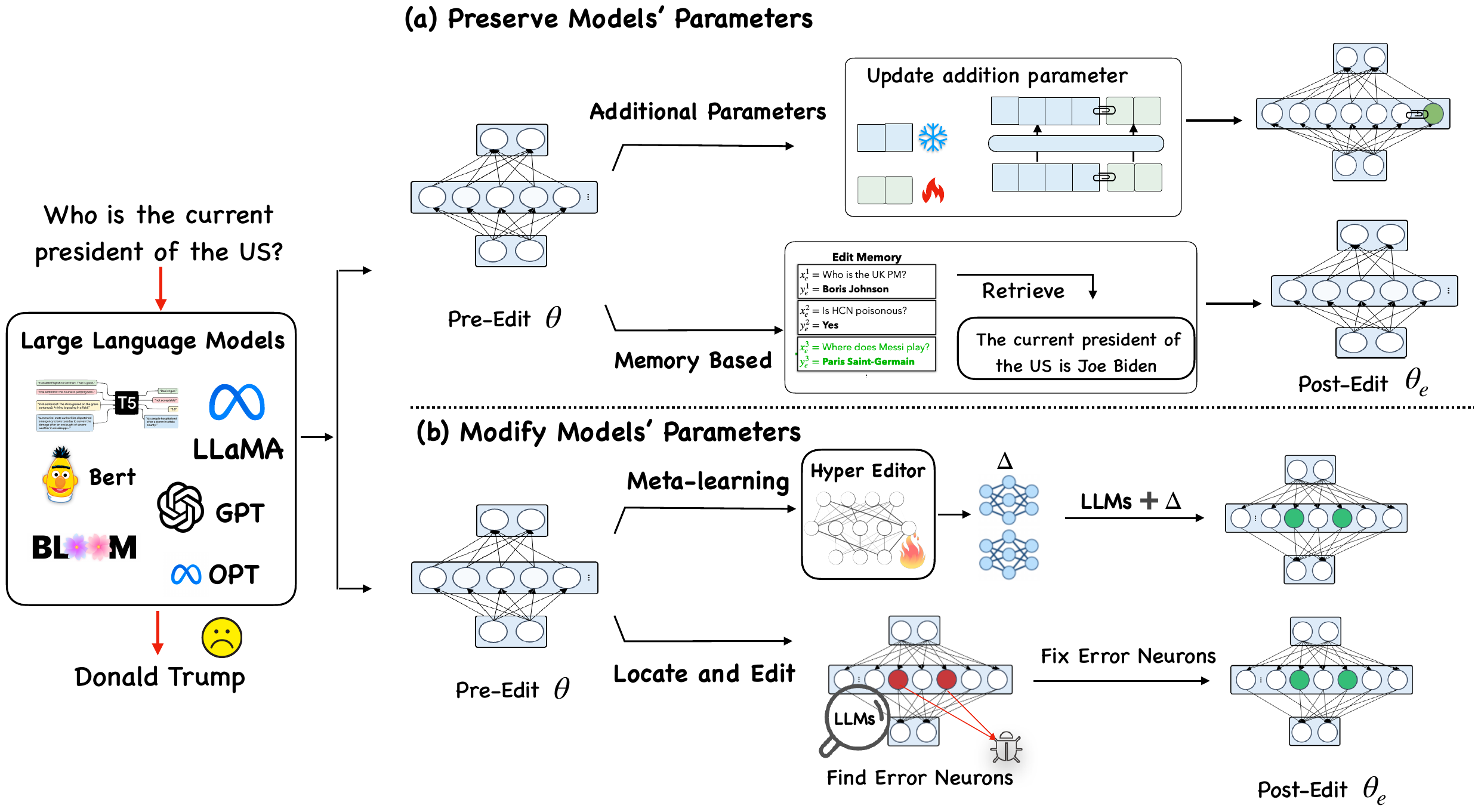}
    \caption{An overview of two paradigms of model editing for LLMs.}
    \label{fig:overview}
\end{figure*}

Currently, numerous works on model editing for LLMs \cite{de-cao-etal-2021-editing,meng2022locating,meng2023massediting,Sinitsin2020Editable,huang2023transformerpatcher} have made strides in various editing tasks and settings. 
As illustrated in Figure~\ref{fig:overview}, these works manipulate the model's output for specific cases by either integrating an auxiliary network with the original unchanged model or altering the model parameters responsible for the undesirable output.
Despite the wide range of model editing techniques present in the literature, a comprehensive comparative analysis, assessing these methods in uniform experimental conditions, is notably lacking.
This absence of direct comparison impairs our ability to discern the relative merits and demerits of each approach, consequently hindering our comprehension of their adaptability across different problem domains.

To confront this issue, the present study endeavors to establish a standard problem definition accompanied by a meticulous appraisal of these methods (\S{\ref{sec:definition}}, \S{\ref{sec:method}}). 
We conduct experiments under regulated conditions, fostering an impartial comparison of their respective strengths and weaknesses (\S{\ref{sec:preliminary}}). 
We initially use two popular model editing datasets, ZsRE ~\cite{levy-etal-2017-zero} and \cf~\cite{meng2022locating}, and two structurally different language models, T5 \cite{DBLP:journals/jmlr/RaffelSRLNMZLL20} (encoder-decoder) and GPT-J \cite{wang2021gpt} (decoder only), as our base models.
We also evaluate the performance of larger models, OPT-13B~\cite{DBLP:journals/corr/abs-2205-01068} and GPT-NEOX-20B~\cite{black2022gptneox20b}.
Beyond basic edit settings, we assess performance for \textbf{batch} and \textbf{sequential} editing. 
While we observe that current methods have demonstrated considerable capacity in factual model editing tasks, we reconsider the current evaluation and create a more encompassing evaluation dataset(\S{\ref{sec:comprehensive}}): \textbf{portability} (robust generalization capabilities), \textbf{locality} (side effect), and \textbf{efficiency} (time and memory usage).
We find current model editing methods are somewhat limited on these levels, thereby constraining their practical application, and deserve more research in the future.
Through systematic evaluation, we aim to provide valuable insights on each model editing technique's effectiveness, aiding researchers in choosing the appropriate method for specific tasks.


\section{Problems Definition}
\label{sec:definition}
Model editing, as elucidated by \citet{Mitchell2022MemoryBasedME}, aims to adjust an initial base model's ($f_{\theta}$, $\theta$ signifies the model's parameters) behavior on the particular edit descriptor $(x_e, y_e)$ \textbf{efficiently} without influencing the model behavior on other samples. The ultimate goal is to create an edited model, denoted $f_{\theta_{e}}$.
Specifically, the basic model $f_{\theta}$ is represented by a function $f: \mathbb{X} \mapsto \mathbb{Y}$ that associates an input $x$ with its corresponding prediction $y$. Given an edit descriptor comprising the edit input $x_e$ and edit label $y_e$ such that $f_{\theta}(x_e) \neq y_e$, the post-edit model $f_{\theta_{e}}$ is designed to produce the expected output, where $f_{\theta_{e}}(x_e) = y_e$.

The model editing process generally impacts the predictions for a broad set of inputs that are closely associated with the edit example. This collection of inputs is called the \textbf{editing scope}. A successful edit should adjust the model's behavior for examples within the editing scope while leaving its performance for out-of-scope examples unaltered:
\begin{equation}
f_{\theta_{e}}(x) = \begin{cases}
y_e & \text{if } x \in I(x_e,y_e) \\
f_{\theta}(x) & \text{if } x \in O(x_e, y_e)
\end{cases}
\end{equation}
The \emph{in-scope} $I(x_e,y_e)$ usually encompasses $x_e$ along with its equivalence neighborhood $N(x_e, y_e)$, which includes related input/output pairs. In contrast, the \emph{out-of-scope} $O(x_e, y_e)$ consists of inputs that are unrelated to the edit example.
The post-edit model $f_e$ should satisfy the following three properties: \textbf{reliability}, \textbf{generalization}, and \textbf{locality}.

\paragraph{Reliability}
Previous works~\cite{huang2023transformerpatcher,de-cao-etal-2021-editing,meng2022locating} define a reliable edit when the post-edit model $f_{\theta_{e}}$ gives the target answer for the case $(x_e,y_e)$ to be edited.
The reliability is measured as the average accuracy on the edit case:
\begin{equation}
\mathbb{E}_{x_{\mathrm{e}}^{\prime}, y_{\mathrm{e}}^{\prime} \sim \left\{\left(x_{\mathrm{e}}, y_{\mathrm{e}}\right)\right\}} \mathbbm{1} \left\{\operatorname{argmax}_y f_{\theta_{e}}\left(y \mid x_{\mathrm{e}}^{\prime}\right)=y_{\mathrm{e}}^{\prime}\right\}
\end{equation}
\paragraph{Generalization}
The post-edit model $f_{\theta_{e}}$ should also edit the equivalent neighbour $N\left(x_{\mathrm{e}}, y_{\mathrm{e}}\right)$ (e.g. rephrased sentences). 
It is evaluated by the average accuracy of the model $f_{\theta_{e}}$ on examples drawn uniformly from the equivalence neighborhood:
\begin{equation}
\mathbb{E}_{x_{\mathrm{e}}^{\prime}, y_{\mathrm{e}}^{\prime} \sim N\left(x_{\mathrm{e}}, y_{\mathrm{e}}\right)} \mathbbm {1} \left\{\operatorname{argmax}_yf_{\theta_{e}}\left(y \mid x_{\mathrm{e}}^{\prime}\right)=y_{\mathrm{e}}^{\prime}\right\}
\end{equation} 

\paragraph{Locality}
also noted as \textbf{Specificity} in some work. Editing should be implemented locally, which means the post-edit model $f_{\theta_{e}}$ should not change the output of the irrelevant examples in the out-of-scope $O(x_e,y_e)$. Hence, the locality is evaluated by the rate at which the
post-edit model $f_{\theta_{e}}$’s predictions are unchanged as the pre-edit $f_\theta$ model:
\begin{equation}
\mathbb{E}_{x_{\mathrm{e}}^{\prime}, y_{\mathrm{e}}^{\prime} \sim O\left(x_{\mathrm{e}}, y_{\mathrm{e}}\right)} \mathbbm {1} \left\{f_{\theta_{e}}\left(y \mid x_{\mathrm{e}}^{\prime}\right)=f_{\theta}\left(y \mid x_{\mathrm{e}}^{\prime}\right) \right\}
\end{equation}

\section{Current Methods}
\label{sec:method}
Current model editing methods for LLMs can be categorized into two main paradigms as shown in Figure~\ref{fig:overview}: modifying the model's parameters or preserving the model's parameters. 
More comparisons can be seen in Table~\ref{tab:conceptual_analysis}.
\subsection{\fontsize{10.5pt}{\baselineskip}\selectfont Methods for Preserving LLMs' Parameters}
\paragraph{Memory-based Model}
This kind of method stores all edit examples explicitly in memory and employs a retriever to extract the most relevant edit facts for each new input to guide the model to generate the edited fact. 
SERAC~\cite{Mitchell2022MemoryBasedME} presents an approach that adopts a distinct \emph{counterfactual model} while leaving the original model unchanged.
Specifically, it employs a \emph{scope classifier} to compute the likelihood of new input falling within the purview of stored edit examples.
If the input matches any cached edit in memory, the counterfactual model's prediction is based on the input and the most probable edit. Otherwise, if the input is out-of-scope for all edits, the original model's prediction is given.
Additionally, recent research demonstrates that LLMs possess robust capabilities for \emph{in-context learning}.
Instead of resorting to an extra model trained with new facts, the model itself can generate outputs corresponding to the provided knowledge given a refined knowledge context as a prompt.
This kind of method edits the language model by prompting the model with the edited fact and retrieved edit demonstrations from the edit memory and includes the following work: MemPrompt~\cite{madaan-etal-2022-memory},IKE~\cite{Zheng2023CanWE} and MeLLo~\cite{zhong2023mquake}.
\paragraph{Additional Parameters}
This paradigm introduces extra trainable parameters within the language models. 
These parameters are trained on a modified knowledge dataset while the original model parameters remain static. 
T-Patcher~\cite{huang2023transformerpatcher} integrates one neuron(patch) for one mistake in the last layer of the Feed-Forward Network (FFN) of the model, which takes effect only when encountering its corresponding mistake.
CaliNET~\cite{dong-etal-2022-calibrating} incorporates several neurons for multiple edit cases.
Differently, GRACE \cite{Hartvigsen2022AgingWG} maintains a discrete codebook as an Adapter, adding and updating elements over time to edit a model’s predictions.
\subsection{\fontsize{10.5pt}{\baselineskip}\selectfont Methods for Modifying LLMs' Parameters}
This paradigm would update part of the parameter $\theta$, it applies an update $\Delta$ matrix to edit the model.

\definecolor{Mycolor1}{HTML}{BAD8F2}
\definecolor{Mycolor2}{HTML}{DDEEFA}
\begin{table*}[ht]
{
\small
\resizebox{\textwidth}{!}{
\begin{tabular}{lcc|c|cc|cc|cc|ccc}
\toprule
\textbf{DataSet}     & \textbf{Model}                                      & \textbf{Metric}   &  \textbf{FT-L}
& \textbf{SERAC} & \textbf{$\text{IKE}$} & \textbf{CaliNet} &\textbf{T-Patcher} & \textbf{KE}         & \textbf{MEND} & \textbf{KN} &  \textbf{ROME} & \textbf{$\text{MEMIT}$}  \\ 
\midrule
\multirow{7}{*}{\textbf{ZsRE}}        & \multirow{3}{*}{T5-XL}                    
& Reliability &20.71  & \colorbox{Mycolor1}{\textbf{99.80}}   &67.00        &    5.17  & 30.52   & 3.00    & \colorbox{Mycolor2}{78.80}  & 22.51   &   -   &  -    \\
                             &                                           
& Generalization & {19.68} &  \colorbox{Mycolor1}{\textbf{99.66}}  &67.11    &   4.81  & 30.53    &   5.40      &   \colorbox{Mycolor2}{89.80}  & 22.70 &   -   &     - \\
                             &                                           
& Locality   &89.01  &  \colorbox{Mycolor2}{98.13} &63.60   &   72.47    & 77.10  &   96.43      &     \colorbox{Mycolor1}{\textbf{98.45}} &  16.43 &    -  &    -   \\ 
\cmidrule{2-13} 
                             & \multirow{3}{*}{GPT-J}                     
& Reliability &54.70 & 90.16 &\colorbox{Mycolor1}{\textbf{99.96}} & 22.72 & 97.12      &   6.60    &  98.15  & 11.34 &  99.18 &    \colorbox{Mycolor2}{99.23}    \\
                             &                                          
& Generalization &49.20 & 89.96  & \colorbox{Mycolor1}{\textbf{99.87}} & 0.12 & 94.95 & 7.80 & \colorbox{Mycolor2}{97.66}  & 9.40 &  94.90  &    87.16  \\
                             &                                           
& Locality   &37.24 & \colorbox{Mycolor1}{\textbf{99.90}} &59.21  &   12.03   & 96.24    &    94.18        &   97.39    & 90.03       &  99.19  &   \colorbox{Mycolor2}{99.62}   \\
\midrule
\multirow{7}{*}{\textbf{\textsc{CounterFact}}} & \multirow{3}{*}{T5-XL}                    
& Reliability &33.57 & \colorbox{Mycolor1}{\textbf{99.89}} &\colorbox{Mycolor2}{97.77} &   7.76     & 80.26    &  1.00            &  81.40       & 47.86    &   -   &  -     \\
                             &                                           
& Generalization &23.54 & \colorbox{Mycolor1}{\textbf{98.71}} & 82.99    &   7.57    & 21.73       &   1.40           &  \colorbox{Mycolor2}{93.40}     & 46.78      &   -   &  -     \\
                             &                                          
& Locality   &72.72 &  \colorbox{Mycolor1}{\textbf{99.93}} &37.76   &  27.75  & 85.09      &    \colorbox{Mycolor2}{96.28}        &  91.58          & 57.10      &    -  &   -   \\ 
\cmidrule{2-13} 
                             & \multicolumn{1}{l}{\multirow{3}{*}{GPT-J}} 
& Reliability &\colorbox{Mycolor2}{99.90} & 99.78  & 99.61 &43.58  & \colorbox{Mycolor1}{\textbf{100.00}}&  13.40 & 73.80& 1.66 & 99.80 & \colorbox{Mycolor2}{99.90}      \\
                             &                   
& Generalization &\colorbox{Mycolor2}{97.53} &  \colorbox{Mycolor1}{\textbf{99.41}} & 72.67    &0.66&   83.98 &  11.00     & 74.20    & 1.38   &  86.63 & 73.13  \\
                             & 
& Locality&1.02 & \colorbox{Mycolor1}{\textbf{98.89}}& 35.57 &2.69&8.37&94.38&93.75&58.28&93.61& \colorbox{Mycolor2}{97.17}   \\ 
\bottomrule
\end{tabular}
}
}
\caption{Results of existing methods on three metrics of the dataset.
The settings for these models and datasets are the same with \citet{meng2022locating}. `-' refers to the results that the methods empirically fail to edit LLMs.}
\label{tab:experimental_analysis}
\end{table*}

\paragraph{Locate-Then-Edit}
This paradigm initially identifies parameters corresponding to specific knowledge and modifies them through direct updates to the target parameters.
The Knowledge Neuron (KN) method~\cite{dai-etal-2022-knowledge} introduces a \emph{knowledge attribution} technique to pinpoint the ``knowledge neuron'' (a key-value pair in the FFN matrix) that embodies the knowledge and then updates these neurons. 
ROME~\cite{meng2022locating} applies causal mediation analysis to locate the editing area. 
Instead of modifying the knowledge neurons in the FFN, ROME alters the entire matrix.
ROME views model editing as the least squares with a linear equality constraint and uses the Lagrange multiplier to solve it.
However, KN and ROME can only edit one factual association at a time.
To this end, MEMIT~\cite{meng2023massediting} expands on the setup of ROME, realizing the situation of synchronous editing for multiple cases.
Based on MEMIT, PMET~\cite{Li2023PMETPM} involves the attention value to get a better performance.
\paragraph{Meta-learning}
Meta-learning methods employ a hyper network to learn the necessary $\Delta$ for editing the LLMs. 
Knowledge Editor (KE)~\cite{de-cao-etal-2021-editing} leverages a hypernetwork (specifically, a bidirectional-LSTM) to predict the weight update for each data point, thereby enabling the constrained optimization of editing target knowledge without disrupting others.
However, this approach falls short when it comes to editing LLMs.
To overcome this limitation, Model Editor Networks with Gradient Decomposition (MEND)~\cite{mitchell2022fast} learns to transform the gradient of fine-tuned language models by employing a low-rank decomposition of gradients, which can be applied to LLMs with better performance.
\section{Preliminary Experiments}
\label{sec:preliminary}

Considering the abundance of studies and datasets centered on factual knowledge, we use it as our primary comparison foundation.
Our initial controlled experiments, conducted using two prominent factual knowledge datasets (Table \ref{tab:experimental_analysis}), facilitate a direct comparison of methods, highlighting their unique strengths and limitations \cite{DBLP:journals/corr/abs-2308-07269}.

\subsection{Experiment Setting}
We use two prominent model editing datasets: ZsRE and \textsc{CounterFact}, with their details available in Appendix~\ref{app:dataset}. 
Previous studies typically used smaller language models (<1B) and demonstrated the effectiveness of current editing methods on smaller models like BERT~\cite{devlin-etal-2019-bert}.
However, whether these methods work for larger models is still unexplored.
Hence, considering the editing task and future developments, we focus on generation-based models and choose larger ones: T5-XL (3B) and GPT-J (6B), representing both encoder-decoder and decoder-only structures. 

We've selected influential works from each method type. 
Alongside existing model editing techniques, we additionally examined the results of fine-tuning, an elementary approach for model updating. 
To avoid the computational cost of retraining all layers, we employed methodology proposed by \citet{meng2022locating}, fine-tuning layers identified by ROME and we denoted it as FT-L. 
This strategy ensures a fair comparison with other direct editing methods, bolstering our analysis's validity.
More details can be found in Appendix \ref{sec:appendix}.
\begin{figure*}
    \centering
    \includegraphics[width=1\textwidth]{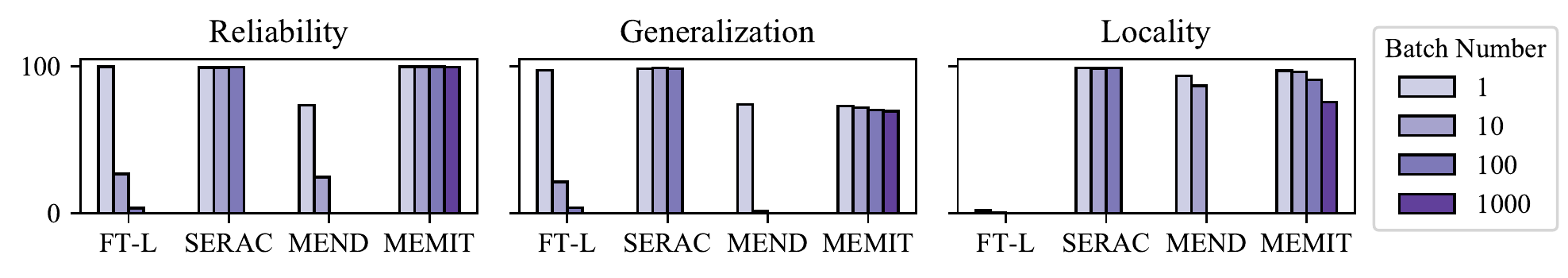}
    \caption{\textbf{Batch Editing} performance against batch number. We test batch numbers in [1,10,100,1000] for MEMIT. Due to the huge memory usage for FT, SERAC and MEND, we didn't test batch 1000 for these methods.}
    \label{fig:batch_edit}
\end{figure*}
\subsection{Experiment Results}
\paragraph{Basic Model}
Table~\ref{tab:experimental_analysis} reveals SERAC and ROME's superior performance on the ZsRE and \textsc{CounterFact} datasets, with SERAC exceeding 90\% on several metrics. 
While MEMIT lacks its generalization, it excels in reliability and locality.
KE, CaliNET, and KN perform poorly, with acceptable performance in smaller models, but mediocrity in larger ones.
MEND performs well on the two datasets, achieving over 80\% in the results on T5, although not as impressive as ROME and SERAC.
The performance of the T-Patcher model fluctuates across different model architectures and sizes.
For instance, it underperforms on T5-XL for the ZsRE dataset, while it performs perfectly on GPT-J.
In the case of the \textsc{CounterFact} dataset, T-Patcher achieves satisfactory reliability and locality on T5 but lacks generalization. 
Conversely, on GPT-J, the model excels in reliability and generalization but underperforms in the locality.
This instability can be attributed to the model architecture since T-Patcher adds a neuron to the final decoder layer for T5; however, the encoder may still retain the original knowledge. 
FT-L performs less impressively than ROME on PLMs, even when modifying the same position. It shows underwhelming performance on the ZsRE dataset but equals ROME in reliability and generalization with the \cf~ dataset on GPT-J. Yet, its low locality score suggests potential impacts on unrelated knowledge areas.
IKE demonstrates good reliability but struggles with locality, as prepended prompts might affect unrelated inputs. Its generalization capability could also improve. 
The in-context learning method may struggle with context mediation failure~\cite{hernandez2023inspecting}, as pre-trained language models may not consistently generate text aligned with the prompt. 

\paragraph{Model Scaling}
We conduct experiments with larger models, testing IKE, ROME, and MEMIT on OPT-13B and GPT-NEOX-20B due to computational constraints.
The results (Table~\ref{tab:basic_larger}) surprisingly show ROME and MEMIT performing well on the GPT-NEOX-20B model but failing on OPT-13B. 
This is due to both methods relying on a \emph{matrix inversion} operation.
However, in the OPT-13B model, the matrix is not \emph{invertible}.
We even empirically find that approximating the solution with least squares yields unsatisfactory results.
We think this is the limitation of ROME and MEMIT as they are based on the strong assumption that matrices are non-degenerate and may not be applied to different models.
MEMIT performs worse due to its reliance on multi-layer matrix computations, and its reliability and generalization declined more than ROME's for larger models.
IKE's performance is affected by the in-context learning ability of the model itself. 
The results of OPT are even worse than the results of GPT-J, which may be attributed to OPT's own in-context learning ability.
Additionally, as the model size increases, its performance in both generalization and locality diminishes.

\begin{figure}[t]
    \hspace*{-0.4cm}
    \includegraphics[width=0.5\textwidth]{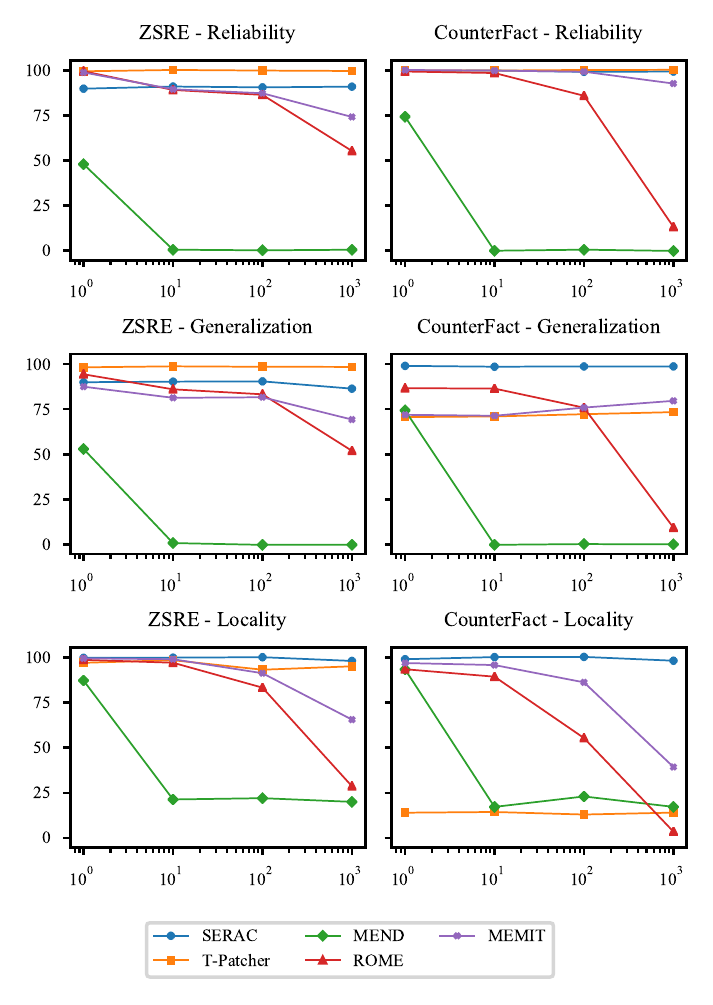}
    \caption{\textbf{Sequential Editing} performance against data stream size (log-scale). }
    \label{fig:sequential_edit}
\end{figure}

\paragraph{Batch Editing}
We conduct further batch editing analysis, given that many studies often limit updates to a few dozen facts or focus only on single-edit cases. 
However, it's often necessary to modify the model with multiple knowledge pieces simultaneously. 
We focused on batch-editing-supportive methods (FT, SERAC, MEND, and MEMIT) and displayed their performance in Figure~\ref{fig:batch_edit}.
\begin{table}[t]
\centering
\resizebox{1.0\columnwidth}{!}{
\begin{tabular}{lccc|ccc}
\toprule
         \multirow{2}{*}{\textbf{Method}}        & \multicolumn{3}{c}{\textbf{ZSRE}}   & \multicolumn{3}{c}{\textbf{\cf}}                                                                 \\ \cmidrule{2-7}
 & Reliability & Generalization & Locality & \multicolumn{1}{c}{Reliability} & \multicolumn{1}{c}{Generalization} & \multicolumn{1}{c}{Locality} \\ \midrule
 \textit{OPT-13B} \\ \midrule
 ROME & \colorbox{Mycolor2}{22.23} & \colorbox{Mycolor1}{6.08} &\colorbox{Mycolor1}{\textbf{99.74}}& \colorbox{Mycolor2}{36.85} & \colorbox{Mycolor2}{2.86} & \colorbox{Mycolor1}{\textbf{95.46}}  \\
MEMIT & 7.95 & 2.87 & \colorbox{Mycolor1}{92.61} & 4.95 & 0.36 & \colorbox{Mycolor2}{93.28}  \\
IKE  &\colorbox{Mycolor1}{\textbf{69.97}} & \colorbox{Mycolor1}{\textbf{69.93}}&64.83  & \colorbox{Mycolor1}{\textbf{49.71}} & \colorbox{Mycolor1}{\textbf{34.98}} &53.08 \\ 
\midrule
 \textit{GPT-NEOX-20B} \\ \midrule
ROME & \colorbox{Mycolor2}{99.34} & \colorbox{Mycolor2}{95.49} & \colorbox{Mycolor1}{\textbf{99.79}} & \colorbox{Mycolor1}{\textbf{99.80}} & \colorbox{Mycolor1}{\textbf{85.45}}  &  \colorbox{Mycolor2}{94.54}     \\
MEMIT & 77.30 & 71.44 & \colorbox{Mycolor2}{99.67} & 87.22 & \colorbox{Mycolor2}{70.26} & \colorbox{Mycolor1}{\textbf{96.48}}  \\
IKE  & \colorbox{Mycolor1}{\textbf{100.00}}  & \colorbox{Mycolor1}{\textbf{99.95}}   &59.69  & \colorbox{Mycolor2}{98.64} &67.67 &43.03 \\ \bottomrule                        
\end{tabular}}
\caption{Current methods' results of current datasets on \textbf{OPT-13B} and \textbf{GPT-NEOX-20B}.}
\label{tab:basic_larger}
\end{table}
Notably, MEMIT supports massive knowledge editing for LLMs, allowing hundreds or even thousands of simultaneous edits with minimal time and memory costs. 
Its performance across reliability and generalization remains robust up to 1000 edits, but locality decreases at this level.
While FT-L, SERAC, and MEND also support batch editing, they require significant memory for handling more cases, exceeding our current capabilities. Thus, we limited tests to 100 edits.
SERAC can conduct batch edits perfectly up to 100 edits.
MEND and FT-L performance in batch edits is not as strong, with the model's performance rapidly declining as the number of edits increases.
\paragraph{Sequential Editing}
Note that the default evaluation procedure is to update a single model knowledge, evaluate the new model, and then roll back the update before repeating the process for each test point.
In practical scenarios, models should retain previous changes while conducting new edits.
Thus, the ability to carry out successive edits is a vital feature for model editing~\cite{huang2023transformerpatcher}. 
We evaluate approaches with strong single-edit performance for sequential editing and report the results in Figure~\ref{fig:sequential_edit}.
Methods that freeze the model's parameters, like SERAC and T-Patcher, generally show stable performance in sequential editing. 
However, those altering the model's parameters struggle. 
ROME performs well up to $n = 10$, then degrades at $n = 100$. 
MEMIT's performance also decreases over 100 edits, but less drastically than ROME.
Similarly, MEND performs well at $n = 1$ but significantly declines at $n = 10$.
As the editing process continues, these models increasingly deviate from their original state, resulting in suboptimal performance.
\section{Comprehensive Study}
\label{sec:comprehensive}
Considering the above points, we contend that previous evaluation metrics may not fully assess model editing capabilities. 
Therefore, we propose more comprehensive evaluations regarding \textbf{portability}, \textbf{locality}, and \textbf{efficiency}.
\subsection{Portability - Robust Generalization}
Several studies evaluate generalization using samples generated through back translation~\cite{de-cao-etal-2021-editing}. However, these paraphrased sentences often involve only minor wording changes and don't reflect substantial factual modifications. 
As stated in \citet{Jacquesrome}, it's crucial to verify if these methods can handle the implications of an edit for realistic applications.
As a result, we introduce a new evaluation metric called \textbf{Portability} to gauge the effectiveness of model editing in transferring knowledge to related content, termed robust generalization. 
Hence we consider three aspects:
(1) \textbf{Subject Replace}: As most rephrased sentences keep subject descriptions but rephrase the relation more, we test generalization by replacing the subject in the question with an alias or synonym. This tests whether the model can generalize the edited attribute to other descriptions of the same subject.
(2) \textbf{Reversed Relation}: When the target of a subject and relation is edited, the attribute of the target entity also changes. We test the model's ability to handle this by filtering for suitable relations such as one-to-one and asking it the reverse question to check if the target entity is also updated.
(3) \textbf{One-hop}: Modified knowledge should be usable by the edited language model for downstream tasks. For example, if we change the answer to the question ``What university did Watts Humphrey attend?'' from ``Trinity College'' to ``University of Michigan'', the model should then answer ``Ann Arbor in Michigan State'' instead of ``Dublin in Ireland'' when asked, ``Which city did Watts Humphrey live in during his university studies?'' 
We thus construct a reasoning dataset to evaluate the post-edit models' abilities to use the edited knowledge.
\definecolor{Mycolor1}{HTML}{BAD8F2}
\definecolor{Mycolor2}{HTML}{DDEEFA}

\begin{table}[t]
\centering
\small
\resizebox{1.0\columnwidth}{!}{
\begin{tabular}{lccc}
\toprule
& \textbf{Subject-} &  \textbf{Reverse-} &  \textbf{One-} \\
\textbf{Method} & \textbf{Replace} &  \textbf{Relation} &  \textbf{hop} \\ \midrule
\textit{GPT-J-6B}             \\ \midrule
FT-L          &72.96            &8.05                   & 1.34 \\
SERAC        &17.79            &1.30                  & 5.53 \\
T-Patcher        &\colorbox{Mycolor1}{\textbf{96.65}}            &33.62                   &3.10  \\
MEND        &  42.45          &  0.00                 & 11.34 \\
ROME        &37.42            &46.42                   & 50.91 \\
MEMIT       &27.73            &\colorbox{Mycolor2}{47.67}  & \colorbox{Mycolor2}{52.74} \\
IKE         &\colorbox{Mycolor2}{88.77}   & \colorbox{Mycolor1}{\textbf{92.96}}   &\colorbox{Mycolor1}{\textbf{55.38}}         \\ \midrule
\textit{GPT-NEOX-20B}             \\ \midrule     
ROME        &\colorbox{Mycolor2}{44.57} & 48.99 & \colorbox{Mycolor2}{51.03}      \\
MEMIT       &30.98   &\colorbox{Mycolor2}{49.19}     &   49.58       \\
IKE         & \colorbox{Mycolor1}{\textbf{85.54}}      & \colorbox{Mycolor1}{\textbf{96.46}}                 &\colorbox{Mycolor1}{\textbf{58.97}}         \\
\bottomrule
\end{tabular}
}
\caption{Portability results on various model editing methods. The example for each assessment type can be found in Table\ref{tab:port-dataset-example} at Appendix~\ref{app:dataset}.}
\vspace{-6mm}
\label{tab:portability}
\end{table}

We incorporate a new part, $P(x_e, y_e)$, into the existing dataset ZsRE, and \textbf{Portability} is calculated as the average accuracy of the edited model $(f_{\theta_e})$ when applied to reasoning examples in $P(x_e, y_e)$:
\begin{equation}
\mathbb{E}_{x_{\mathrm{e}}^{\prime}, y_{\mathrm{e}}^{\prime} \sim P\left(x_{\mathrm{e}}, y_{\mathrm{e}}\right)} \mathbbm {1} \left\{\operatorname{argmax}_y f_{\theta_e}\left(y \mid x_{\mathrm{e}}^{\prime}\right)=y_{\mathrm{e}}^{\prime}\right\}
\end{equation}

\paragraph{Dataset Construction} 
As to the one-hop dataset, in the original edit, we alter the answer from $o$ to $o^*$ for a subject $s$. We then prompt the model to generate a linked triple $(o^*, r^*, o^{'*})$. 
Subsequently, GPT-4 creates a question and answer based on this triple and $s$.
Notably, if the model can answer this new question, it would imply that it has pre-existing knowledge of the triple $(o^*, r^*, o^{'*})$. 
We filter out unknown triples by asking the model to predict $o^{'*}$ from $o^*$ and $r^*$. If successful, it's inferred the model has prior knowledge.
Finally, \textbf{Human evaluators} verify the triple's accuracy and the question's fluency. 
Additional details, such as the demonstrations we used and other parts of dataset construction, can be found in the Appendix \ref{app:dataset}.
\paragraph{Results} 
We conduct experiments based on the newly proposed evaluation metric and datasets, presenting the results in Table~\ref{tab:portability}. 
As demonstrated by the Table, the performance of current model editing methods regarding portability is somewhat suboptimal. 
SERAC, despite showing impeccable results on previous metrics, scores less than 20\% accuracy across all three portability aspects.
The bottleneck of SERAC lies in the accuracy of the classifier and the capabilities of the additional model. 
As to the \emph{subject replace} scenario, including SERAC, MEND, ROME, and MEMIT, can only adapt to a specific subject entity expression but cannot generalize to the concept of the subject entity.
However, FT-L, IKE, and T-patcher demonstrate great performance when facing the substituted subject.
Regarding the \emph{reversed relation}, our results indicate that current editing methods mainly edit one-direction relations, with IKE as the notable exception, achieving over 90\% on both GPT-J and GPT-NEOX-20B.  
Other methods alter the subject entities' attributes while leaving the object entity unaffected.
In the \emph{one-hop} reasoning setting, most of the editing methods struggle to transfer the altered knowledge to related facts.
Unexpectedly, ROME, MEMIT, and IKE exhibit relatively commendable performance on portability (exceeding 50\%). 
They are capable of not only editing the original cases but also modifying facts correlated with them in some respect.
To summarize, IKE exhibits relatively good performance across the three scenarios in our evaluations.
However, it is clear that current model editing techniques continue to face challenges in managing the ramifications of an edit - that is, ensuring that changes to knowledge are coherently and consistently reflected in related contexts. This area, indeed, calls for further investigation and innovation in future research.



\subsection{Locality - Side Effect of Model Editing}
\begin{table}[t]
\centering
\small
\resizebox{1.0\columnwidth}{!}{
\begin{tabular}{lccc}
\toprule
& \textbf{Other-} &  \textbf{Distract-} &  \textbf{Other-} \\
\textbf{Method} & \textbf{Attribution} &  \textbf{Neighbor} &  \textbf{Task} \\ \midrule
FT-L         &12.88            &9.48                   &49.56  \\
MEND        & 73.50           &32.96                   &48.86  \\
SERAC        &\colorbox{Mycolor1}{\textbf{99.50}} &39.18      &74.84  \\
T-Patcher    &91.51     &17.56       &\colorbox{Mycolor2}{75.03}  \\
ROME        &78.94    &50.35                   &52.12  \\
MEMIT       &\colorbox{Mycolor2}{86.78}  &\colorbox{Mycolor2}{60.47}  &74.62  \\
IKE         &84.13   &\colorbox{Mycolor1}{\textbf{66.04}} & \colorbox{Mycolor1}{\textbf{75.33}}   \\
\bottomrule
\end{tabular}
}
\caption{Locality results on various model editing methods for GPT-J. Examples of each type can be seen in Tabel~\ref{tab:loc-dataset-example} at Appendix~\ref{app:dataset}.}
\vspace{-6mm}
\label{tab:locality}
\end{table}
In the preceding section, \textsc{CounterFact} and ZsRE evaluate model editing's locality from different perspectives. 
\textsc{CounterFact} employs triples from the same distribution as the target knowledge, while ZsRE utilizes questions from the distinct Natural Questions dataset.
Notably, some methods, such as T-Patcher, exhibit differing performances on these two datasets. This highlights that the impact of model editing on the language model is multifaceted, necessitating a thorough and comprehensive evaluation to fully appreciate its effects.
To thoroughly examine the potential side effects of model editing, we propose evaluations at three different levels:
(1) \textbf{Other Relations}: Although \citet{meng2022locating} introduced the concept of \emph{essence}, they did not explicitly evaluate it. We argue that other attributes of the subject that have been updated should remain unchanged after editing.
(2) \textbf{Distract Neighbourhood}: \citet{HoelscherObermaier2023DetectingEF} find that if we concatenate the edited cases before other unrelated input, the model tends to be swayed by the edited fact and continue to produce results aligned with the edited cases.
(3) \textbf{Other Tasks}:  Building upon Skill Neuron's assertion~\cite{wang-etal-2022-finding-skill} that feed-forward networks in large language models (LLMs) possess task-specific knowledge capabilities, we introduce a new challenge to assess whether model editing might negatively impact performance on other tasks.
Construction of the dataset details can be found in Appendix~\ref{sec:locality_cons}.
\paragraph{Results}
Table~\ref{tab:locality} presents our results. Notably, current editing methods excel in the \emph{other attributions} aspect, indicating that they only modify the target characteristic without affecting other attributes. However, they generally perform poorly in \emph{Distract-Neighbor} settings, as reflected in the performance drop compared to the results in Table~\ref{tab:experimental_analysis}. 
An exception is IKE, whose performance remains relatively stable due to the fact that it inherently requires the edited fact to be concatenated before the input. 
As for the commonsense reasoning tasks, parameter-preserving methods largely maintain their performance on other tasks. Conversely, methods that alter parameters tend to negatively influence performance, with the exception of MEMIT.
Despite changing parameters, MEMIT maintains strong performance in commonsense tasks, demonstrating its commendable locality.

\begin{table}
    \centering
    \small
    \begin{tabular}{l r r}
        \toprule
        Editor & \cf  & ZsRE \\
        \midrule
        FT-L & 35.94s & 58.86s \\
        SERAC & 5.31s &  6.51s  \\
        CaliNet & \colorbox{Mycolor2}{1.88s} &  \colorbox{Mycolor2}{1.93s} \\
        T-Patcher & 1864.74s & 1825.15s  \\
        KE & 2.20s &   2.21s \\
        MEND & \colorbox{Mycolor1}{\textbf{0.51s}} &  \colorbox{Mycolor1}{\textbf{0.52s}}   \\
        KN & 225.43s &  173.57s  \\
        ROME & 147.2s & 183.0s \\
        MEMIT & 143.2s & 145.6s  \\
        \bottomrule
    \end{tabular}
    \caption{\textbf{Wall clock time} for each edit method conducting 10 edits on GPT-J using one 2$\times$V100 (32G). The calculation of this time involves measuring the duration from providing the edited case to obtaining the post-edited model.}
    \label{tab:efficient}
\end{table}

\subsection{Efficiency}
\begin{figure}
    \hspace*{-0.25cm}
    \includegraphics[width=0.46 \textwidth]{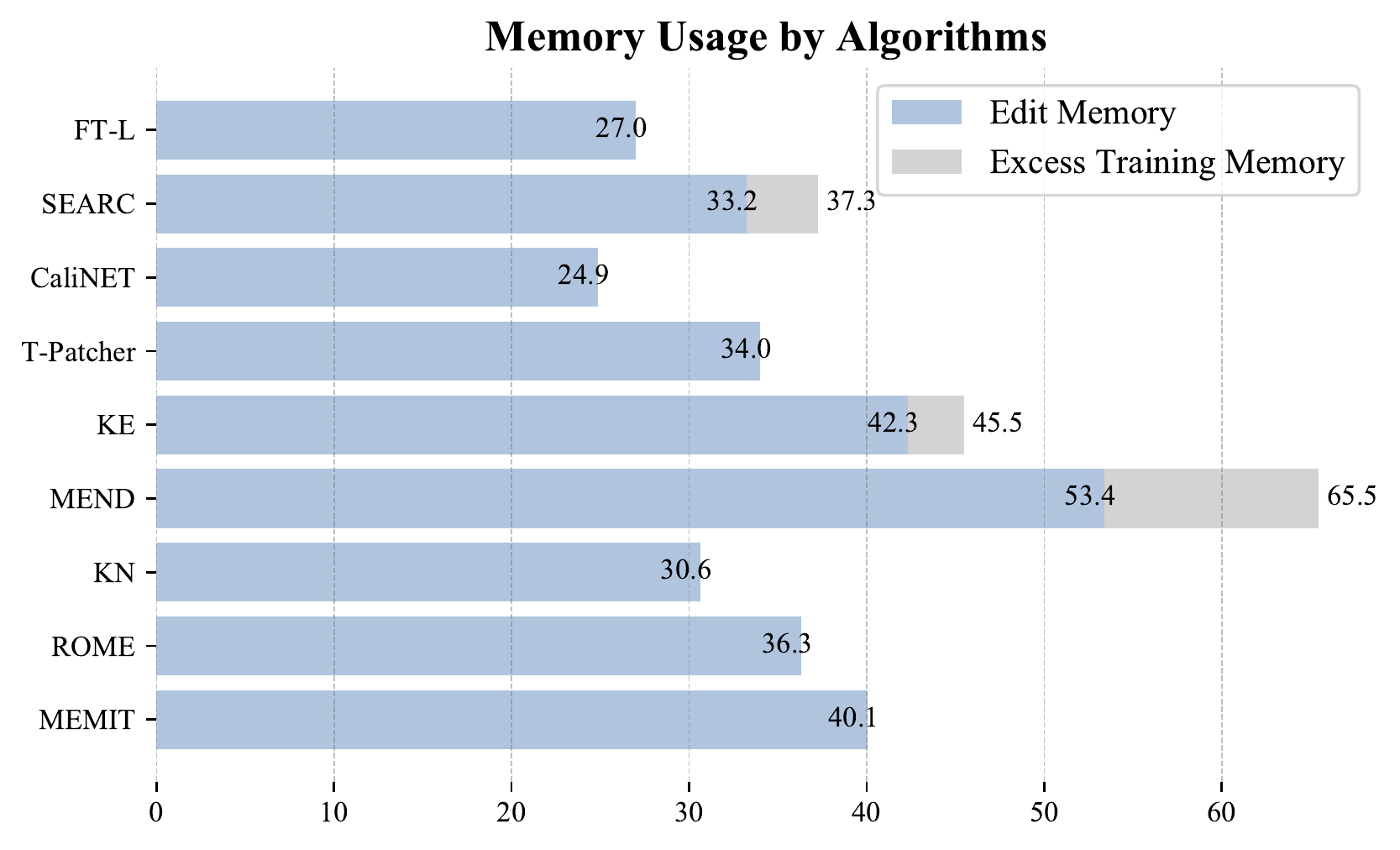}
    \caption{\textbf{GPU VRAM consumption during training and editing} for different model editing methods. 
    }
    \label{fig:memory_usage}
\end{figure}

Model editing should minimize the \textbf{time} and \textbf{memory} required for conducting edits without compromising the model's performance.
\paragraph{Time Analysis}
Table~\ref{tab:efficient} illustrates the time required for different model editing techniques from providing the edited case to obtaining the post-edited model.
We observe that once the hyper-network is trained, KE and MEND perform the editing process at a considerably fast pace.
Likewise, SERAC can also quickly edit knowledge, completing the process in about 5 seconds, given a trained classifier and counterfact model.
However, these methods necessitate \textbf{hours-to-days} of additional training and an extra dataset. 
In our experiments, training MEND on the ZsRE dataset took over 7 hours, and training SERAC required over 36 hours on 3$\times$ V100. 
On the other hand, ROME and MEMIT necessitate a pre-computation of the covariance statistics for the Wikitext.
However, this computation is time-consuming and can potentially take \textbf{hours-to-days} to complete.
In comparison, other methods such as KN, CaliNET, and T-Patcher may be faster since they do not require any pre-computation or pre-training.
However, KN and CaliNET's performance on larger models is unsatisfactory, and T-Patcher is the slowest due to the need for individual neuron training for each corresponding mistake.
Considering the time aspect, there is a need for a model editing method that is more time-friendly.
\paragraph{Memory Analysis}
Figure~\ref{fig:memory_usage} exhibits the memory VRAM usage for each model editing method. From this figure, we observe that the majority of the methods consume a similar amount of memory, with the exception of MEND, which requires more than 60GB for training. 
Methods that introduce extra training, such as MEND and SERAC lead to additional computational overhead, hence the significant increase in memory consumption.
\section{Relationship with Relevant Works}
\subsection{Knowledge in LLMs}
Several model editing approaches aim to discern how knowledge stored in PLMs precisely and directly alters the model's parameters.
There is existing work that examines the principles that govern how PLMs store knowledge~\cite{geva-etal-2021-transformer,geva-etal-2022-transformer,haviv-etal-2023-understanding,Hao2021SelfAttentionAI,hernandez2023inspecting,DBLP:journals/corr/abs-2305-08732,cao2023retentive,lamparth2023analyzing,DBLP:journals/corr/abs-2301-10405,li2023unveiling,DBLP:journals/corr/abs-2308-13198,ju2023klob}, which contribute to the model editing process.
Moreover, some model editing techniques bear resemblance to knowledge augmentation~\cite{DBLP:conf/acl/ZhangHLJSL19,lewis2020retrieval,zhang2022greaselm,yasunaga-etal-2021-qa,Yao2022KformerKI,DBLP:journals/corr/abs-2308-03188} approaches, as updating the model's knowledge can also be considered as instilling knowledge into the model.

\subsection{Lifelong Learning and Unlearning}
Model editing, encompassing lifelong learning and unlearning, allows adaptive addition, modification, and removal of knowledge. 
Continual learning~\cite{Biesialska2020ContinualLL}, which improves model adaptability across tasks and domains, has shown effectiveness in model editing in PLMs~\cite{Zhu2020ModifyingMI}.
Moreover, it's vital for models to forget sensitive knowledge, aligning with machine unlearning concepts~\cite{Hase2023DoesLI, Wu2022PUMAPU,Tarun2021FastYE,DBLP:journals/corr/abs-2303-07345}.


\subsection{Security and Privacy for LLMs}
Past studies \cite{Carlini2020ExtractingTD,shen2023chatgpt} show that LLMs can produce unreliable or personal samples from certain prompts.
The task of erasing potentially harmful and private information stored in large language models (LLMs) is vital to enhance the privacy and security of LLM-based applications \cite{DBLP:journals/corr/abs-2304-10436}. 
Model editing, which can suppress harmful language generation \cite{geva-etal-2022-transformer,DBLP:journals/corr/abs-2308-08090}, could help address these concerns.


\section{Conclusion}
We systematically analyze methods for editing large language models (LLMs). 
We aim to help researchers better understand existing editing techniques by examining their features, strengths, and limitations. Our analysis shows much room for improvement, especially in terms of portability, locality, and efficiency. Improved LLM editing could help better align them with the changing needs and values of users. We hope that our work spurs progress on open issues and further research.

\section*{Acknowledgment}
We would like to express gratitude to the anonymous reviewers for their kind comments. 
This work was supported by the National Natural Science Foundation of China (No.62206246), Zhejiang Provincial Natural Science Foundation of China (No. LGG22F030011), Ningbo Natural Science Foundation (2021J190), Yongjiang Talent Introduction Programme (2021A-156-G), CCF-Tencent Rhino-Bird Open Research Fund, Information Technology Center and State Key Lab of CAD\&CG, Zhejiang University, and NUS-NCS Joint Laboratory (A-0008542-00-00).

\section*{Limitations}
There remain several aspects of model editing that are not covered in this paper.
\paragraph{Model Scale \& Architecture}
Due to computational resource constraints, we have only calculated the results for models up to 20B in size here.
Meanwhile, many model editing methods treat the FFN of the model as key-value pairs.
Whether these methods are effective for models with different architectures, such as Llama, remains to be explored.
\paragraph{Editing Scope}
Notably, the application of model editing goes beyond mere factual contexts, underscoring its vast potential.
Elements such as personality, emotions, opinions, and beliefs also fall within the scope of model editing. 
While these aspects have been somewhat explored, they remain relatively uncharted territories and thus are not detailed in this paper.
Furthermore, multilingual editing \cite{Xu2022LanguageAC,wang2023crosslingual,Wu2023EvaKELLMAN} represents an essential research direction that warrants future attention and exploration.
There are also some editing works that can deal with computer vision tasks such as ENN~\cite{Sinitsin2020Editable} and \citet{ilharco2023editing}.
\paragraph{Editing Setting}
In our paper, the comprehensive study~\ref{sec:comprehensive} mainly evaluated the method's performance on one edit.
During the time of our work, \citet{zhong2023mquake} proposed a multi-hop reasoning setting that explored current editing methods' generalization performance for multiple edits simultaneously.
We leave this multiple-edit evaluation for the future.
Besides, this work focused on changing the model's result to reflect specific facts.
\citet{cohen2023evaluating} propose a benchmark for knowledge injection and knowledge update.
However, erasing the knowledge or information stored in LLMs \cite{belrose2023leace,geva-etal-2022-transformer,ishibashi2023knowledge} is also an important direction for investigating.

\paragraph{Editing Black-Box LLMs}
Meanwhile, models like ChatGPT and GPT-4 exhibit remarkable performance on a wide range of natural language tasks but are only accessible through APIs. 
This raises an important question: How can we edit these ``black-box'' models that also tend to produce undesirable outputs during downstream usage?
Presently, there are some works that utilize in-context learning \cite{DBLP:journals/corr/abs-2305-01651} and prompt-based methods \cite{DBLP:conf/emnlp/MurtyMLR22} to modify these models. 
They precede each example with a textual prompt that specifies the adaptation target, which shows promise as a technique for model editing.

\section*{Ethic Consideration}
Model editing pertains to the methods used to alter the behavior of pre-trained models. However, it's essential to bear in mind that ill-intentioned model editing could lead the model to generate harmful or inappropriate outputs. Therefore, ensuring safe and responsible practices in model editing is of paramount importance. The application of such techniques should be guided by ethical considerations, and there should be safeguards to prevent misuse and the production of harmful results.
\textbf{All our data has been carefully checked by humans, and any malicious editing or offensive content has been removed}.

\bibliography{anthology,custom}

\begin{thebibliography}{74}
\expandafter\ifx\csname natexlab\endcsname\relax\def\natexlab#1{#1}\fi

\bibitem[{Anil et~al.(2023)Anil, Dai, Firat, Johnson, Lepikhin, Passos,
  Shakeri, Taropa, Bailey, Chen et~al.}]{anil2023palm}
Rohan Anil, Andrew~M Dai, Orhan Firat, Melvin Johnson, Dmitry Lepikhin,
  Alexandre Passos, Siamak Shakeri, Emanuel Taropa, Paige Bailey, Zhifeng Chen,
  et~al. 2023.
\newblock Palm 2 technical report.
\newblock \emph{arXiv preprint arXiv:2305.10403}.

\bibitem[{Belrose et~al.(2023)Belrose, Schneider-Joseph, Ravfogel, Cotterell,
  Raff, and Biderman}]{belrose2023leace}
Nora Belrose, David Schneider-Joseph, Shauli Ravfogel, Ryan Cotterell, Edward
  Raff, and Stella Biderman. 2023.
\newblock \href {http://arxiv.org/abs/2306.03819} {Leace: Perfect linear
  concept erasure in closed form}.

\bibitem[{Biesialska et~al.(2020)Biesialska, Biesialska, and
  Costa-juss{\`a}}]{Biesialska2020ContinualLL}
Magdalena Biesialska, Katarzyna Biesialska, and Marta~Ruiz Costa-juss{\`a}.
  2020.
\newblock Continual lifelong learning in natural language processing: A survey.
\newblock \emph{ArXiv}, abs/2012.09823.

\bibitem[{Bisk et~al.(2020)Bisk, Zellers, Bras, Gao, and Choi}]{bisk2020piqa}
Yonatan Bisk, Rowan Zellers, Ronan~Le Bras, Jianfeng Gao, and Yejin Choi. 2020.
\newblock \href {https://ojs.aaai.org/index.php/AAAI/article/view/6239}
  {{PIQA:} reasoning about physical commonsense in natural language}.
\newblock In \emph{The Thirty-Fourth {AAAI} Conference on Artificial
  Intelligence, {AAAI} 2020, The Thirty-Second Innovative Applications of
  Artificial Intelligence Conference, {IAAI} 2020, The Tenth {AAAI} Symposium
  on Educational Advances in Artificial Intelligence, {EAAI} 2020, New York,
  NY, USA, February 7-12, 2020}, pages 7432--7439. {AAAI} Press.

\bibitem[{Black et~al.(2022)Black, Biderman, Hallahan, Anthony, Gao, Golding,
  He, Leahy, McDonell, Phang, Pieler, Prashanth, Purohit, Reynolds, Tow, Wang,
  and Weinbach}]{black2022gptneox20b}
Sid Black, Stella Biderman, Eric Hallahan, Quentin Anthony, Leo Gao, Laurence
  Golding, Horace He, Connor Leahy, Kyle McDonell, Jason Phang, Michael Pieler,
  USVSN~Sai Prashanth, Shivanshu Purohit, Laria Reynolds, Jonathan Tow, Ben
  Wang, and Samuel Weinbach. 2022.
\newblock \href {http://arxiv.org/abs/2204.06745} {Gpt-neox-20b: An open-source
  autoregressive language model}.

\bibitem[{Brown et~al.(2020)Brown, Mann, Ryder, Subbiah, Kaplan, Dhariwal,
  Neelakantan, Shyam, Sastry, Askell, Agarwal, Herbert{-}Voss, Krueger,
  Henighan, Child, Ramesh, Ziegler, Wu, Winter, Hesse, Chen, Sigler, Litwin,
  Gray, Chess, Clark, Berner, McCandlish, Radford, Sutskever, and
  Amodei}]{DBLP:conf/nips/BrownMRSKDNSSAA20}
Tom~B. Brown, Benjamin Mann, Nick Ryder, Melanie Subbiah, Jared Kaplan,
  Prafulla Dhariwal, Arvind Neelakantan, Pranav Shyam, Girish Sastry, Amanda
  Askell, Sandhini Agarwal, Ariel Herbert{-}Voss, Gretchen Krueger, Tom
  Henighan, Rewon Child, Aditya Ramesh, Daniel~M. Ziegler, Jeffrey Wu, Clemens
  Winter, Christopher Hesse, Mark Chen, Eric Sigler, Mateusz Litwin, Scott
  Gray, Benjamin Chess, Jack Clark, Christopher Berner, Sam McCandlish, Alec
  Radford, Ilya Sutskever, and Dario Amodei. 2020.
\newblock \href
  {https://proceedings.neurips.cc/paper/2020/hash/1457c0d6bfcb4967418bfb8ac142f64a-Abstract.html}
  {Language models are few-shot learners}.
\newblock In \emph{Advances in Neural Information Processing Systems 33: Annual
  Conference on Neural Information Processing Systems 2020, NeurIPS 2020,
  December 6-12, 2020, virtual}.

\bibitem[{Cao et~al.(2023)Cao, Tang, Lin, Han, Chen, Wang, and
  Sun}]{cao2023retentive}
Boxi Cao, Qiaoyu Tang, Hongyu Lin, Xianpei Han, Jiawei Chen, Tianshu Wang, and
  Le~Sun. 2023.
\newblock Retentive or forgetful? diving into the knowledge memorizing
  mechanism of language models.
\newblock \emph{arXiv preprint arXiv:2305.09144}.

\bibitem[{Carlini et~al.(2020)Carlini, Tram{\`e}r, Wallace, Jagielski,
  Herbert-Voss, Lee, Roberts, Brown, Song, Erlingsson, Oprea, and
  Raffel}]{Carlini2020ExtractingTD}
Nicholas Carlini, Florian Tram{\`e}r, Eric Wallace, Matthew Jagielski, Ariel
  Herbert-Voss, Katherine Lee, Adam Roberts, Tom~B. Brown, Dawn~Xiaodong Song,
  {\'U}lfar Erlingsson, Alina Oprea, and Colin Raffel. 2020.
\newblock Extracting training data from large language models.
\newblock In \emph{USENIX Security Symposium}.

\bibitem[{Chen et~al.(2023)Chen, Cao, Chen, Liu, and
  Zhao}]{DBLP:journals/corr/abs-2308-13198}
Yuheng Chen, Pengfei Cao, Yubo Chen, Kang Liu, and Jun Zhao. 2023.
\newblock \href {https://doi.org/10.48550/arXiv.2308.13198} {Journey to the
  center of the knowledge neurons: Discoveries of language-independent
  knowledge neurons and degenerate knowledge neurons}.
\newblock \emph{CoRR}, abs/2308.13198.

\bibitem[{Cheng et~al.(2023)Cheng, Zhang, Tian, Dai, Xiong, Guo, and
  Chen}]{DBLP:journals/corr/abs-2301-10405}
Siyuan Cheng, Ningyu Zhang, Bozhong Tian, Zelin Dai, Feiyu Xiong, Wei Guo, and
  Huajun Chen. 2023.
\newblock \href {https://doi.org/10.48550/arXiv.2301.10405} {Editing language
  model-based knowledge graph embeddings}.
\newblock \emph{CoRR}, abs/2301.10405.

\bibitem[{Cohen et~al.(2023)Cohen, Biran, Yoran, Globerson, and
  Geva}]{cohen2023evaluating}
Roi Cohen, Eden Biran, Ori Yoran, Amir Globerson, and Mor Geva. 2023.
\newblock \href {http://arxiv.org/abs/2307.12976} {Evaluating the ripple
  effects of knowledge editing in language models}.

\bibitem[{Dai et~al.(2022)Dai, Dong, Hao, Sui, Chang, and
  Wei}]{dai-etal-2022-knowledge}
Damai Dai, Li~Dong, Yaru Hao, Zhifang Sui, Baobao Chang, and Furu Wei. 2022.
\newblock \href {https://doi.org/10.18653/v1/2022.acl-long.581} {Knowledge
  neurons in pretrained transformers}.
\newblock In \emph{Proceedings of the 60th Annual Meeting of the Association
  for Computational Linguistics (Volume 1: Long Papers)}, pages 8493--8502,
  Dublin, Ireland. Association for Computational Linguistics.

\bibitem[{De~Cao et~al.(2021)De~Cao, Aziz, and
  Titov}]{de-cao-etal-2021-editing}
Nicola De~Cao, Wilker Aziz, and Ivan Titov. 2021.
\newblock \href {https://doi.org/10.18653/v1/2021.emnlp-main.522} {Editing
  factual knowledge in language models}.
\newblock In \emph{Proceedings of the 2021 Conference on Empirical Methods in
  Natural Language Processing}, pages 6491--6506, Online and Punta Cana,
  Dominican Republic. Association for Computational Linguistics.

\bibitem[{Devlin et~al.(2019)Devlin, Chang, Lee, and
  Toutanova}]{devlin-etal-2019-bert}
Jacob Devlin, Ming-Wei Chang, Kenton Lee, and Kristina Toutanova. 2019.
\newblock \href {https://doi.org/10.18653/v1/N19-1423} {{BERT}: Pre-training of
  deep bidirectional transformers for language understanding}.
\newblock In \emph{Proceedings of the 2019 Conference of the North {A}merican
  Chapter of the Association for Computational Linguistics: Human Language
  Technologies, Volume 1 (Long and Short Papers)}, pages 4171--4186,
  Minneapolis, Minnesota. Association for Computational Linguistics.

\bibitem[{Dong et~al.(2022)Dong, Dai, Song, Xu, Sui, and
  Li}]{dong-etal-2022-calibrating}
Qingxiu Dong, Damai Dai, Yifan Song, Jingjing Xu, Zhifang Sui, and Lei Li.
  2022.
\newblock \href {https://aclanthology.org/2022.findings-emnlp.438} {Calibrating
  factual knowledge in pretrained language models}.
\newblock In \emph{Findings of the Association for Computational Linguistics:
  EMNLP 2022}, pages 5937--5947, Abu Dhabi, United Arab Emirates. Association
  for Computational Linguistics.

\bibitem[{Gandikota et~al.(2023)Gandikota, Materzynska, Fiotto{-}Kaufman, and
  Bau}]{DBLP:journals/corr/abs-2303-07345}
Rohit Gandikota, Joanna Materzynska, Jaden Fiotto{-}Kaufman, and David Bau.
  2023.
\newblock \href {https://doi.org/10.48550/arXiv.2303.07345} {Erasing concepts
  from diffusion models}.
\newblock \emph{CoRR}, abs/2303.07345.

\bibitem[{Geva et~al.(2022)Geva, Caciularu, Wang, and
  Goldberg}]{geva-etal-2022-transformer}
Mor Geva, Avi Caciularu, Kevin Wang, and Yoav Goldberg. 2022.
\newblock \href {https://aclanthology.org/2022.emnlp-main.3} {Transformer
  feed-forward layers build predictions by promoting concepts in the vocabulary
  space}.
\newblock In \emph{Proceedings of the 2022 Conference on Empirical Methods in
  Natural Language Processing}, pages 30--45, Abu Dhabi, United Arab Emirates.
  Association for Computational Linguistics.

\bibitem[{Geva et~al.(2021)Geva, Schuster, Berant, and
  Levy}]{geva-etal-2021-transformer}
Mor Geva, Roei Schuster, Jonathan Berant, and Omer Levy. 2021.
\newblock \href {https://doi.org/10.18653/v1/2021.emnlp-main.446} {Transformer
  feed-forward layers are key-value memories}.
\newblock In \emph{Proceedings of the 2021 Conference on Empirical Methods in
  Natural Language Processing}, pages 5484--5495, Online and Punta Cana,
  Dominican Republic. Association for Computational Linguistics.

\bibitem[{Hao et~al.(2021)Hao, Dong, Wei, and Xu}]{Hao2021SelfAttentionAI}
Y.~Hao, Li~Dong, Furu Wei, and Ke~Xu. 2021.
\newblock Self-attention attribution: Interpreting information interactions
  inside transformer.
\newblock In \emph{Proc. of AAAI}.

\bibitem[{Hartvigsen et~al.(2022)Hartvigsen, Sankaranarayanan, Palangi, Kim,
  and Ghassemi}]{Hartvigsen2022AgingWG}
Thomas Hartvigsen, Swami Sankaranarayanan, Hamid Palangi, Yoon Kim, and Marzyeh
  Ghassemi. 2022.
\newblock \href {https://api.semanticscholar.org/CorpusID:253735429} {Aging
  with grace: Lifelong model editing with discrete key-value adaptors}.
\newblock \emph{ArXiv}, abs/2211.11031.

\bibitem[{Hase et~al.(2023)Hase, Bansal, Kim, and
  Ghandeharioun}]{Hase2023DoesLI}
Peter Hase, Mohit Bansal, Been Kim, and Asma Ghandeharioun. 2023.
\newblock Does localization inform editing? surprising differences in
  causality-based localization vs. knowledge editing in language models.
\newblock \emph{ArXiv}, abs/2301.04213.

\bibitem[{Haviv et~al.(2023)Haviv, Cohen, Gidron, Schuster, Goldberg, and
  Geva}]{haviv-etal-2023-understanding}
Adi Haviv, Ido Cohen, Jacob Gidron, Roei Schuster, Yoav Goldberg, and Mor Geva.
  2023.
\newblock \href {https://aclanthology.org/2023.eacl-main.19} {Understanding
  transformer memorization recall through idioms}.
\newblock In \emph{Proceedings of the 17th Conference of the European Chapter
  of the Association for Computational Linguistics}, pages 248--264, Dubrovnik,
  Croatia. Association for Computational Linguistics.

\bibitem[{Hernandez et~al.(2023)Hernandez, Li, and
  Andreas}]{hernandez2023inspecting}
Evan Hernandez, Belinda~Z. Li, and Jacob Andreas. 2023.
\newblock \href {http://arxiv.org/abs/2304.00740} {Inspecting and editing
  knowledge representations in language models}.

\bibitem[{Hoelscher-Obermaier et~al.(2023{\natexlab{a}})Hoelscher-Obermaier,
  Persson, Kran, Konstas, and Barez}]{HoelscherObermaier2023DetectingEF}
J.~Hoelscher-Obermaier, Julia Persson, Esben Kran, Ioannis Konstas, and Fazl
  Barez. 2023{\natexlab{a}}.
\newblock Detecting edit failures in large language models: An improved
  specificity benchmark.
\newblock In \emph{ACL Findings}.

\bibitem[{Hoelscher-Obermaier et~al.(2023{\natexlab{b}})Hoelscher-Obermaier,
  Persson, Kran, Konstas, and Barez}]{jason2023detecting}
Jason Hoelscher-Obermaier, Julia Persson, Esben Kran, Ionnis Konstas, and Fazl
  Barez. 2023{\natexlab{b}}.
\newblock Detecting edit failures in large language models: An improved
  specificity benchmark.
\newblock In \emph{Findings of ACL}. Association for Computational Linguistics.

\bibitem[{Hu et~al.(2023)Hu, Li, Zheng, Liu, Hu, and
  Zhang}]{DBLP:journals/corr/abs-2308-08090}
Xinshuo Hu, Dongfang Li, Zihao Zheng, Zhenyu Liu, Baotian Hu, and Min Zhang.
  2023.
\newblock \href {https://doi.org/10.48550/arXiv.2308.08090} {Separate the wheat
  from the chaff: Model deficiency unlearning via parameter-efficient module
  operation}.
\newblock \emph{CoRR}, abs/2308.08090.

\bibitem[{Huang et~al.(2023)Huang, Shen, Zhang, Zhou, Rong, and
  Xiong}]{huang2023transformerpatcher}
Zeyu Huang, Yikang Shen, Xiaofeng Zhang, Jie Zhou, Wenge Rong, and Zhang Xiong.
  2023.
\newblock \href {https://openreview.net/forum?id=4oYUGeGBPm}
  {Transformer-patcher: One mistake worth one neuron}.
\newblock In \emph{The Eleventh International Conference on Learning
  Representations}.

\bibitem[{Ilharco et~al.(2023)Ilharco, Ribeiro, Wortsman, Schmidt, Hajishirzi,
  and Farhadi}]{ilharco2023editing}
Gabriel Ilharco, Marco~Tulio Ribeiro, Mitchell Wortsman, Ludwig Schmidt,
  Hannaneh Hajishirzi, and Ali Farhadi. 2023.
\newblock \href {https://openreview.net/forum?id=6t0Kwf8-jrj} {Editing models
  with task arithmetic}.
\newblock In \emph{The Eleventh International Conference on Learning
  Representations}.

\bibitem[{Ishibashi and Shimodaira(2023)}]{ishibashi2023knowledge}
Yoichi Ishibashi and Hidetoshi Shimodaira. 2023.
\newblock Knowledge sanitization of large language models.
\newblock \emph{arXiv preprint arXiv:2309.11852}.

\bibitem[{{Jacques Thibodeau}(2022)}]{Jacquesrome}
{Jacques Thibodeau}. 2022.
\newblock But is it really in rome? an investigation of the rome model editing
  technique.

\bibitem[{Ju and Zhang(2023)}]{ju2023klob}
Yiming Ju and Zheng Zhang. 2023.
\newblock Klob: a benchmark for assessing knowledge locating methods in
  language models.
\newblock \emph{arXiv preprint arXiv:2309.16535}.

\bibitem[{Kingma and Ba(2014)}]{kingma2014adam}
Diederik~P Kingma and Jimmy Ba. 2014.
\newblock Adam: A method for stochastic optimization.
\newblock \emph{arXiv preprint arXiv:1412.6980}.

\bibitem[{Kwiatkowski et~al.(2019)Kwiatkowski, Palomaki, Redfield, Collins,
  Parikh, Alberti, Epstein, Polosukhin, Devlin, Lee, Toutanova, Jones, Kelcey,
  Chang, Dai, Uszkoreit, Le, and Petrov}]{kwiatkowski-etal-2019-natural}
Tom Kwiatkowski, Jennimaria Palomaki, Olivia Redfield, Michael Collins, Ankur
  Parikh, Chris Alberti, Danielle Epstein, Illia Polosukhin, Jacob Devlin,
  Kenton Lee, Kristina Toutanova, Llion Jones, Matthew Kelcey, Ming-Wei Chang,
  Andrew~M. Dai, Jakob Uszkoreit, Quoc Le, and Slav Petrov. 2019.
\newblock \href {https://doi.org/10.1162/tacl_a_00276} {Natural questions: A
  benchmark for question answering research}.
\newblock \emph{Transactions of the Association for Computational Linguistics},
  7:452--466.

\bibitem[{Lamparth and Reuel(2023)}]{lamparth2023analyzing}
Max Lamparth and Anka Reuel. 2023.
\newblock Analyzing and editing inner mechanisms of backdoored language models.
\newblock \emph{arXiv preprint arXiv:2302.12461}.

\bibitem[{Levy et~al.(2017)Levy, Seo, Choi, and
  Zettlemoyer}]{levy-etal-2017-zero}
Omer Levy, Minjoon Seo, Eunsol Choi, and Luke Zettlemoyer. 2017.
\newblock \href {https://doi.org/10.18653/v1/K17-1034} {Zero-shot relation
  extraction via reading comprehension}.
\newblock In \emph{Proceedings of the 21st Conference on Computational Natural
  Language Learning ({C}o{NLL} 2017)}, pages 333--342, Vancouver, Canada.
  Association for Computational Linguistics.

\bibitem[{Lewis et~al.(2020)Lewis, Perez, Piktus, Petroni, Karpukhin, Goyal,
  K{\"u}ttler, Lewis, Yih, Rockt{\"a}schel et~al.}]{lewis2020retrieval}
Patrick Lewis, Ethan Perez, Aleksandra Piktus, Fabio Petroni, Vladimir
  Karpukhin, Naman Goyal, Heinrich K{\"u}ttler, Mike Lewis, Wen-tau Yih, Tim
  Rockt{\"a}schel, et~al. 2020.
\newblock Retrieval-augmented generation for knowledge-intensive nlp tasks.
\newblock \emph{Advances in Neural Information Processing Systems},
  33:9459--9474.

\bibitem[{Li et~al.(2023{\natexlab{a}})Li, Li, Song, Yang, Ma, and
  Yu}]{Li2023PMETPM}
Xiaopeng Li, Shasha Li, Shezheng Song, Jing Yang, Jun Ma, and Jie Yu.
  2023{\natexlab{a}}.
\newblock \href {https://api.semanticscholar.org/CorpusID:261030625} {Pmet:
  Precise model editing in a transformer}.

\bibitem[{Li et~al.(2023{\natexlab{b}})Li, Zhang, Yao, Wang, Chen, and
  Chen}]{li2023unveiling}
Zhoubo Li, Ningyu Zhang, Yunzhi Yao, Mengru Wang, Xi~Chen, and Huajun Chen.
  2023{\natexlab{b}}.
\newblock Unveiling the pitfalls of knowledge editing for large language
  models.
\newblock \emph{arXiv preprint arXiv:2310.02129}.

\bibitem[{Madaan et~al.(2022)Madaan, Tandon, Clark, and
  Yang}]{madaan-etal-2022-memory}
Aman Madaan, Niket Tandon, Peter Clark, and Yiming Yang. 2022.
\newblock \href {https://aclanthology.org/2022.emnlp-main.183} {Memory-assisted
  prompt editing to improve {GPT}-3 after deployment}.
\newblock In \emph{Proceedings of the 2022 Conference on Empirical Methods in
  Natural Language Processing}, pages 2833--2861, Abu Dhabi, United Arab
  Emirates. Association for Computational Linguistics.

\bibitem[{Meng et~al.(2022)Meng, Bau, Andonian, and
  Belinkov}]{meng2022locating}
Kevin Meng, David Bau, Alex Andonian, and Yonatan Belinkov. 2022.
\newblock \href {https://openreview.net/forum?id=-h6WAS6eE4} {Locating and
  editing factual associations in {GPT}}.
\newblock \emph{Advances in Neural Information Processing Systems}, 36.

\bibitem[{Meng et~al.(2023)Meng, Sharma, Andonian, Belinkov, and
  Bau}]{meng2023massediting}
Kevin Meng, Arnab~Sen Sharma, Alex~J Andonian, Yonatan Belinkov, and David Bau.
  2023.
\newblock \href {https://openreview.net/forum?id=MkbcAHIYgyS} {Mass-editing
  memory in a transformer}.
\newblock In \emph{The Eleventh International Conference on Learning
  Representations}.

\bibitem[{Mitchell et~al.(2022{\natexlab{a}})Mitchell, Lin, Bosselut, Finn, and
  Manning}]{mitchell2022fast}
Eric Mitchell, Charles Lin, Antoine Bosselut, Chelsea Finn, and Christopher~D
  Manning. 2022{\natexlab{a}}.
\newblock \href {https://openreview.net/forum?id=0DcZxeWfOPt} {Fast model
  editing at scale}.
\newblock In \emph{International Conference on Learning Representations}.

\bibitem[{Mitchell et~al.(2022{\natexlab{b}})Mitchell, Lin, Bosselut, Manning,
  and Finn}]{Mitchell2022MemoryBasedME}
Eric Mitchell, Charles Lin, Antoine Bosselut, Christopher~D. Manning, and
  Chelsea Finn. 2022{\natexlab{b}}.
\newblock Memory-based model editing at scale.
\newblock In \emph{International Conference on Machine Learning}.

\bibitem[{Murty et~al.(2022)Murty, Manning, Lundberg, and
  Ribeiro}]{DBLP:conf/emnlp/MurtyMLR22}
Shikhar Murty, Christopher~D. Manning, Scott~M. Lundberg, and Marco~T{\'{u}}lio
  Ribeiro. 2022.
\newblock \href {https://aclanthology.org/2022.emnlp-main.797} {Fixing model
  bugs with natural language patches}.
\newblock In \emph{Proceedings of the 2022 Conference on Empirical Methods in
  Natural Language Processing, {EMNLP} 2022, Abu Dhabi, United Arab Emirates,
  December 7-11, 2022}, pages 11600--11613. Association for Computational
  Linguistics.

\bibitem[{Onoe et~al.(2023)Onoe, Zhang, Padmanabhan, Durrett, and
  Choi}]{DBLP:journals/corr/abs-2305-01651}
Yasumasa Onoe, Michael J.~Q. Zhang, Shankar Padmanabhan, Greg Durrett, and
  Eunsol Choi. 2023.
\newblock \href {https://doi.org/10.48550/arXiv.2305.01651} {Can lms learn new
  entities from descriptions? challenges in propagating injected knowledge}.
\newblock \emph{CoRR}, abs/2305.01651.

\bibitem[{OpenAI(2023)}]{DBLP:journals/corr/abs-2303-08774}
OpenAI. 2023.
\newblock \href {https://doi.org/10.48550/arXiv.2303.08774} {{GPT-4} technical
  report}.
\newblock \emph{CoRR}, abs/2303.08774.

\bibitem[{Pan et~al.(2023)Pan, Saxon, Xu, Nathani, Wang, and
  Wang}]{DBLP:journals/corr/abs-2308-03188}
Liangming Pan, Michael Saxon, Wenda Xu, Deepak Nathani, Xinyi Wang, and
  William~Yang Wang. 2023.
\newblock \href {https://doi.org/10.48550/arXiv.2308.03188} {Automatically
  correcting large language models: Surveying the landscape of diverse
  self-correction strategies}.
\newblock \emph{CoRR}, abs/2308.03188.

\bibitem[{Qiao et~al.(2022)Qiao, Ou, Zhang, Chen, Yao, Deng, Tan, Huang, and
  Chen}]{DBLP:journals/corr/abs-2212-09597}
Shuofei Qiao, Yixin Ou, Ningyu Zhang, Xiang Chen, Yunzhi Yao, Shumin Deng,
  Chuanqi Tan, Fei Huang, and Huajun Chen. 2022.
\newblock \href {https://doi.org/10.48550/arXiv.2212.09597} {Reasoning with
  language model prompting: {A} survey}.
\newblock \emph{CoRR}, abs/2212.09597.

\bibitem[{Raffel et~al.(2020{\natexlab{a}})Raffel, Shazeer, Roberts, Lee,
  Narang, Matena, Zhou, Li, and Liu}]{DBLP:journals/jmlr/RaffelSRLNMZLL20}
Colin Raffel, Noam Shazeer, Adam Roberts, Katherine Lee, Sharan Narang, Michael
  Matena, Yanqi Zhou, Wei Li, and Peter~J. Liu. 2020{\natexlab{a}}.
\newblock \href {http://jmlr.org/papers/v21/20-074.html} {Exploring the limits
  of transfer learning with a unified text-to-text transformer}.
\newblock \emph{J. Mach. Learn. Res.}, 21:140:1--140:67.

\bibitem[{Raffel et~al.(2020{\natexlab{b}})Raffel, Shazeer, Roberts, Lee,
  Narang, Matena, Zhou, Li, and Liu}]{JMLR:v21:20-074}
Colin Raffel, Noam Shazeer, Adam Roberts, Katherine Lee, Sharan Narang, Michael
  Matena, Yanqi Zhou, Wei Li, and Peter~J. Liu. 2020{\natexlab{b}}.
\newblock \href {http://jmlr.org/papers/v21/20-074.html} {Exploring the limits
  of transfer learning with a unified text-to-text transformer}.
\newblock \emph{Journal of Machine Learning Research}, 21(140):1--67.

\bibitem[{Sanh et~al.(2019)Sanh, Debut, Chaumond, and
  Wolf}]{DBLP:journals/corr/abs-1910-01108}
Victor Sanh, Lysandre Debut, Julien Chaumond, and Thomas Wolf. 2019.
\newblock \href {http://arxiv.org/abs/1910.01108} {Distilbert, a distilled
  version of {BERT:} smaller, faster, cheaper and lighter}.
\newblock \emph{CoRR}, abs/1910.01108.

\bibitem[{Shen et~al.(2023)Shen, Chen, Backes, and Zhang}]{shen2023chatgpt}
Xinyue Shen, Zeyuan Chen, Michael Backes, and Yang Zhang. 2023.
\newblock \href {http://arxiv.org/abs/2304.08979} {In chatgpt we trust?
  measuring and characterizing the reliability of chatgpt}.

\bibitem[{Sinitsin et~al.(2020)Sinitsin, Plokhotnyuk, Pyrkin, Popov, and
  Babenko}]{Sinitsin2020Editable}
Anton Sinitsin, Vsevolod Plokhotnyuk, Dmitry Pyrkin, Sergei Popov, and Artem
  Babenko. 2020.
\newblock \href {https://openreview.net/forum?id=HJedXaEtvS} {Editable neural
  networks}.
\newblock In \emph{International Conference on Learning Representations}.

\bibitem[{Sun et~al.(2023)Sun, Zhang, Deng, Cheng, and
  Huang}]{DBLP:journals/corr/abs-2304-10436}
Hao Sun, Zhexin Zhang, Jiawen Deng, Jiale Cheng, and Minlie Huang. 2023.
\newblock \href {https://doi.org/10.48550/arXiv.2304.10436} {Safety assessment
  of chinese large language models}.
\newblock \emph{CoRR}, abs/2304.10436.

\bibitem[{Tarun et~al.(2021)Tarun, Chundawat, Mandal, and
  Kankanhalli}]{Tarun2021FastYE}
Ayush~K Tarun, Vikram~S Chundawat, Murari Mandal, and Mohan~S. Kankanhalli.
  2021.
\newblock Fast yet effective machine unlearning.
\newblock \emph{IEEE transactions on neural networks and learning systems}, PP.

\bibitem[{Touvron et~al.(2023)Touvron, Lavril, Izacard, Martinet, Lachaux,
  Lacroix, Rozi{\`{e}}re, Goyal, Hambro, Azhar, Rodriguez, Joulin, Grave, and
  Lample}]{DBLP:journals/corr/abs-2302-13971}
Hugo Touvron, Thibaut Lavril, Gautier Izacard, Xavier Martinet, Marie{-}Anne
  Lachaux, Timoth{\'{e}}e Lacroix, Baptiste Rozi{\`{e}}re, Naman Goyal, Eric
  Hambro, Faisal Azhar, Aur{\'{e}}lien Rodriguez, Armand Joulin, Edouard Grave,
  and Guillaume Lample. 2023.
\newblock \href {https://doi.org/10.48550/arXiv.2302.13971} {Llama: Open and
  efficient foundation language models}.
\newblock \emph{CoRR}, abs/2302.13971.

\bibitem[{Wang and Komatsuzaki(2021{\natexlab{a}})}]{wang2021gpt}
Ben Wang and Aran Komatsuzaki. 2021{\natexlab{a}}.
\newblock Gpt-j-6b: A 6 billion parameter autoregressive language model.

\bibitem[{Wang and Komatsuzaki(2021{\natexlab{b}})}]{gpt-j}
Ben Wang and Aran Komatsuzaki. 2021{\natexlab{b}}.
\newblock {GPT-J-6B: A 6 Billion Parameter Autoregressive Language Model}.
\newblock \url{https://github.com/kingoflolz/mesh-transformer-jax}.

\bibitem[{Wang et~al.(2023{\natexlab{a}})Wang, Liang, Sun, Cao, and
  Xu}]{wang2023crosslingual}
Jiaan Wang, Yunlong Liang, Zengkui Sun, Yuxuan Cao, and Jiarong Xu.
  2023{\natexlab{a}}.
\newblock \href {http://arxiv.org/abs/2309.08952} {Cross-lingual knowledge
  editing in large language models}.

\bibitem[{Wang et~al.(2023{\natexlab{b}})Wang, Zhang, Xie, Yao, Tian, Wang, Xi,
  Cheng, Liu, Zheng, and Chen}]{DBLP:journals/corr/abs-2308-07269}
Peng Wang, Ningyu Zhang, Xin Xie, Yunzhi Yao, Bozhong Tian, Mengru Wang, Zekun
  Xi, Siyuan Cheng, Kangwei Liu, Guozhou Zheng, and Huajun Chen.
  2023{\natexlab{b}}.
\newblock \href {https://doi.org/10.48550/arXiv.2308.07269} {Easyedit: An
  easy-to-use knowledge editing framework for large language models}.
\newblock \emph{CoRR}, abs/2308.07269.

\bibitem[{Wang et~al.(2022)Wang, Wen, Zhang, Hou, Liu, and
  Li}]{wang-etal-2022-finding-skill}
Xiaozhi Wang, Kaiyue Wen, Zhengyan Zhang, Lei Hou, Zhiyuan Liu, and Juanzi Li.
  2022.
\newblock \href {https://aclanthology.org/2022.emnlp-main.765} {Finding skill
  neurons in pre-trained transformer-based language models}.
\newblock In \emph{Proceedings of the 2022 Conference on Empirical Methods in
  Natural Language Processing}, pages 11132--11152, Abu Dhabi, United Arab
  Emirates. Association for Computational Linguistics.

\bibitem[{Wu et~al.(2022)Wu, Hashemi, and Srinivasa}]{Wu2022PUMAPU}
Ga~Wu, Masoud Hashemi, and Christopher Srinivasa. 2022.
\newblock Puma: Performance unchanged model augmentation for training data
  removal.
\newblock In \emph{AAAI Conference on Artificial Intelligence}.

\bibitem[{Wu et~al.(2023)Wu, Peng, Chen, Su, and Sun}]{Wu2023EvaKELLMAN}
Suhang Wu, Minlong Peng, Yue Chen, Jinsong Su, and Mingming Sun. 2023.
\newblock \href {https://api.semanticscholar.org/CorpusID:261049822}
  {Eva-kellm: A new benchmark for evaluating knowledge editing of llms}.

\bibitem[{Xu et~al.(2022)Xu, Hou, and Che}]{Xu2022LanguageAC}
Yang Xu, Yutai Hou, and Wanxiang Che. 2022.
\newblock Language anisotropic cross-lingual model editing.
\newblock \emph{ArXiv}, abs/2205.12677.

\bibitem[{Yao et~al.(2022)Yao, Huang, Zhang, Dong, Wei, and
  Chen}]{Yao2022KformerKI}
Yunzhi Yao, Shaohan Huang, Ningyu Zhang, Li~Dong, Furu Wei, and Huajun Chen.
  2022.
\newblock Kformer: Knowledge injection in transformer feed-forward layers.
\newblock In \emph{Natural Language Processing and Chinese Computing}.

\bibitem[{Yao et~al.(2023)Yao, Wang, Mao, Tan, Huang, Chen, and
  Zhang}]{DBLP:journals/corr/abs-2305-08732}
Yunzhi Yao, Peng Wang, Shengyu Mao, Chuanqi Tan, Fei Huang, Huajun Chen, and
  Ningyu Zhang. 2023.
\newblock \href {https://doi.org/10.48550/arXiv.2305.08732} {Knowledge
  rumination for pre-trained language models}.
\newblock \emph{CoRR}, abs/2305.08732.

\bibitem[{Yasunaga et~al.(2021)Yasunaga, Ren, Bosselut, Liang, and
  Leskovec}]{yasunaga-etal-2021-qa}
Michihiro Yasunaga, Hongyu Ren, Antoine Bosselut, Percy Liang, and Jure
  Leskovec. 2021.
\newblock \href {https://doi.org/10.18653/v1/2021.naacl-main.45} {{QA}-{GNN}:
  Reasoning with language models and knowledge graphs for question answering}.
\newblock In \emph{Proceedings of the 2021 Conference of the North American
  Chapter of the Association for Computational Linguistics: Human Language
  Technologies}, pages 535--546, Online. Association for Computational
  Linguistics.

\bibitem[{Zhang et~al.(2022{\natexlab{a}})Zhang, Roller, Goyal, Artetxe, Chen,
  Chen, Dewan, Diab, Li, Lin, Mihaylov, Ott, Shleifer, Shuster, Simig, Koura,
  Sridhar, Wang, and Zettlemoyer}]{DBLP:journals/corr/abs-2205-01068}
Susan Zhang, Stephen Roller, Naman Goyal, Mikel Artetxe, Moya Chen, Shuohui
  Chen, Christopher Dewan, Mona~T. Diab, Xian Li, Xi~Victoria Lin, Todor
  Mihaylov, Myle Ott, Sam Shleifer, Kurt Shuster, Daniel Simig, Punit~Singh
  Koura, Anjali Sridhar, Tianlu Wang, and Luke Zettlemoyer. 2022{\natexlab{a}}.
\newblock \href {https://doi.org/10.48550/arXiv.2205.01068} {{OPT:} open
  pre-trained transformer language models}.
\newblock \emph{CoRR}, abs/2205.01068.

\bibitem[{Zhang et~al.(2022{\natexlab{b}})Zhang, Bosselut, Yasunaga, Ren,
  Liang, Manning, and Leskovec}]{zhang2022greaselm}
Xikun Zhang, Antoine Bosselut, Michihiro Yasunaga, Hongyu Ren, Percy Liang,
  Christopher~D Manning, and Jure Leskovec. 2022{\natexlab{b}}.
\newblock \href {https://openreview.net/forum?id=41e9o6cQPj} {Grease{LM}: Graph
  {REAS}oning enhanced language models}.
\newblock In \emph{International Conference on Learning Representations}.

\bibitem[{Zhang et~al.(2019)Zhang, Han, Liu, Jiang, Sun, and
  Liu}]{DBLP:conf/acl/ZhangHLJSL19}
Zhengyan Zhang, Xu~Han, Zhiyuan Liu, Xin Jiang, Maosong Sun, and Qun Liu. 2019.
\newblock \href {https://doi.org/10.18653/v1/p19-1139} {{ERNIE:} enhanced
  language representation with informative entities}.
\newblock In \emph{Proceedings of the 57th Conference of the Association for
  Computational Linguistics, {ACL} 2019, Florence, Italy, July 28- August 2,
  2019, Volume 1: Long Papers}, pages 1441--1451. Association for Computational
  Linguistics.

\bibitem[{Zhao et~al.(2023)Zhao, Zhou, Li, Tang, Wang, Hou, Min, Zhang, Zhang,
  Dong, Du, Yang, Chen, Chen, Jiang, Ren, Li, Tang, Liu, Liu, Nie, and
  Wen}]{DBLP:journals/corr/abs-2303-18223}
Wayne~Xin Zhao, Kun Zhou, Junyi Li, Tianyi Tang, Xiaolei Wang, Yupeng Hou,
  Yingqian Min, Beichen Zhang, Junjie Zhang, Zican Dong, Yifan Du, Chen Yang,
  Yushuo Chen, Zhipeng Chen, Jinhao Jiang, Ruiyang Ren, Yifan Li, Xinyu Tang,
  Zikang Liu, Peiyu Liu, Jian{-}Yun Nie, and Ji{-}Rong Wen. 2023.
\newblock \href {https://doi.org/10.48550/arXiv.2303.18223} {A survey of large
  language models}.
\newblock \emph{CoRR}, abs/2303.18223.

\bibitem[{Zheng et~al.(2023)Zheng, Li, Dong, Fan, Wu, Xu, and
  Chang}]{Zheng2023CanWE}
Ce~Zheng, Lei Li, Qingxiu Dong, Yixuan Fan, Zhiyong Wu, Jingjing Xu, and Baobao
  Chang. 2023.
\newblock Can we edit factual knowledge by in-context learning?
\newblock \emph{ArXiv}, abs/2305.12740.

\bibitem[{Zhong et~al.(2023)Zhong, Wu, Manning, Potts, and
  Chen}]{zhong2023mquake}
Zexuan Zhong, Zhengxuan Wu, Christopher~D. Manning, Christopher Potts, and
  Danqi Chen. 2023.
\newblock \href {http://arxiv.org/abs/2305.14795} {Mquake: Assessing knowledge
  editing in language models via multi-hop questions}.

\bibitem[{Zhu et~al.(2020)Zhu, Rawat, Zaheer, Bhojanapalli, Li, Yu, and
  Kumar}]{Zhu2020ModifyingMI}
Chen Zhu, Ankit~Singh Rawat, Manzil Zaheer, Srinadh Bhojanapalli, Daliang Li,
  Felix~X. Yu, and Sanjiv Kumar. 2020.
\newblock Modifying memories in transformer models.
\newblock \emph{ArXiv}, abs/2012.00363.

\end{thebibliography}
\bibliographystyle{acl_natbib}

\appendix

\section{Implementing Details}
\begin{table*}[!tp]
\centering
\resizebox{\textwidth}{!}{
\begin{tabular}{lllccccc}
\toprule
&&\textbf{Approach}  & \textbf{\makecell{Additional\\Training}} & \textbf{\makecell{Edit\\Type}} & \textbf{\makecell{Batch\\Edit}} & \textbf{\makecell{Edit\\Area}} & \textbf{\makecell{Editor \\ Parameters}}  \\ \hline
\multirow{4}{*}{\makecell{Preserve \\ Parameters}} & \multirow{2}{*}{Memory-based}   
& SERAC  & YES  & Fact\&Sentiment     & YES        & External Model    & $Model_{cf}+Model_{Classifier}$  \\ 
 & &  IKE & NO   & Fact\&Sentiment  & NO  & Input    & NONE   \\ \cline{2-8}
 & \multirow{2}{*}{Additional-Parameters}    & CaliNET  & NO   & Fact       & YES   & FFN     & $N*neuron$   \\
 &   &  T-Patcher & NO   & Fact    & NO  & FFN     & $N*neuron$    \\ 
 \hline
\multirow{5}{*}{\makecell{Modify \\ Parameters}}  
& \multirow{2}{*}{Meta-learning}  
& KE       &    YES     & Fact      & YES     & FFN    &   $Model_{hyper}+L*mlp$     \\
 &     &  MEND     & YES    & Fact    & YES    & FFN   & $Model_{hyper}+L*mlp$       \\
 \cline{2-8}
& \multirow{3}{*}{Locate and Edit}  
& KN & NO        & Fact    & NO       & FFN   & $L*neuron$     \\
&    & ROME    & NO     & Fact  & NO  & FFN   & $mlp_{proj}$      \\
 &  &  MEMIT   & NO & Fact  & YES & FFN   & $L*mlp_{proj}$    \\ 
\bottomrule
\end{tabular}
}
\caption{Comparisons between several existing model editing approaches.
``Additional Training'' refers to whether the methods need training before conducting specific edits. 
``Edit Type'' refers to the format the method can edit. ``Batch Edit'' refers to editing multiple target knowledge simultaneously. ``Editor Area'' refers to the specific region of the LLMs that the methods aim to modify. FFN demonstrates the feed-forward module.
``Editor Parameters'' refers to the parameters that need to be updated for editing. $L$ denotes the number of layers to update. $mlp$ is FFN and $mlp_{proj}$ is the second linear layer in FFN.
$neurons$ denotes the key-value pair in FFN.
$N$ represents the quantity of $neuron$ to be updated within a single layer.
}
\label{tab:conceptual_analysis}
\end{table*}
\label{sec:appendix}
Since the ZsRE dataset adopts the NQ dataset to evaluate the locality, here, we use a T5-XL model~\cite{JMLR:v21:20-074} finetuned on the NQ dataset as the baseline model. 
As to the GPT-J~\cite{gpt-j}, we use the original pre-trained version to test the locality's zero-shot results.
As several original implementations do not support both architectures, we have re-implemented them to accommodate both models. 
We re-implemented some original implementations to support both models.
However, our empirical findings suggest that ROME and MEMIT are only suitable for decoder-only models like GPT-J, so we have not reported results for T5-XL. 
\paragraph{FT}
For basic Fine-Tuning (FT), we follow \citet{meng2023massediting} re-implementation in their study, which uses Adam \cite{kingma2014adam} with early stopping to minimize $-log\mathbb{P}_{G'}[o*\mid p]$, changing only $mlp_{proj}$ weights at selected layer 21. 
For both models, all hyperparameters follow default settings. 
To ensure fairness in the experiments, we always use the unconstrained fine-tuning approach.

\paragraph{KE}
\citet{de-cao-etal-2021-editing} develops an LSTM sequence model, which employs gradient information to predict the rank-1 weight alterations in $G$. 
Given that the official code doesn't facilitate GPT-J, we resort to using the re-implemented version provided by \citet{mitchell2022fast} in their research. 
To foster an equitable comparison across both zsRE and \textsc{CounterFact} tasks, we have taken additional steps to train KE-zsRE and KE-CF models.
The hyperparameters employed for training have been sourced from the default configurations provided.
During testing, KE presents a scaling factor to tweak the norm of the weight update, for which we adhere to the default value of 1.0.

\paragraph{CaliNET}
\citet{dong-etal-2022-calibrating} enriches the FFN by incorporating extra parameters aimed at knowledge editing, comprised of a number of calibration memory slots.
In order to adapt CaliNET to the task at hand, we retain the same architecture as used in the Feed-Forward Network (FFN), albeit with a reduced intermediate dimension denoted as $d$. 
This adaptation allows us to effectively apply CaliNET while managing the model's complexity and computational requirements.
Regarding hyperparameters, we implement adjustments to the FFN within the final two layers of GPT-J, while all other configurations remain consistent with the default settings.

\paragraph{MEND} ~\citet{mitchell2022fast} develop an efficient method for locally editing language models using just a single input-output pair. Essentially, MEND employs a technique to manipulate the gradient of fine-tuned language models which leverages a low-rank decomposition of gradients.
The hyperparameters follow default settings, with the exception of several experiments conducted on GPT-J. 
Specifically, we adjust the optimizer from Adam to AdamW.

\paragraph{SERAC}
~\citet{Mitchell2022MemoryBasedME} presents a method for model editing, named MEME (Memory-Based Model Editing), which stores edits in an explicit memory and learns to reason over them to adjust the base model’s predictions as needed.
The system uses an explicit cache of user-provided edit descriptors (arbitrary utterances for language models), alongside a small auxiliary \emph{scope classifier} and \emph{counterfactual model}. The scope classifier determines if the input falls within the scope of any cached items, and if so, the counterfactual model uses the input and the most relevant edit example to make a prediction.

In alignment with the original paper, we use publicly available Huggingface implementations and checkpoints for all experiments. For the SERAC \emph{scope classifier} model, we adopt \texttt{distilbert-base-cased} \cite{DBLP:journals/corr/abs-1910-01108} across all models and experimental settings. For the counterfactual model, we employ \texttt{T5-small} \cite{JMLR:v21:20-074} for the T5-XL implementation and \texttt{architext/gptj-162M} (available at here\footnote{\url{https://huggingface.co/architext/gptj-162M}}) for the GPT-J implementation. 
All hyperparameters for training and test-time inference are derived from default configurations.

Similar to T-Patcher, in auto-regressive model (like GPT-J) training, we only consider loss at the output positions.

\paragraph{KN}

For Knowledge Neuron \cite{dai-etal-2022-knowledge},  we follow \citet{meng2023massediting} re-implementation in their study. 
The method begins by identifying neurons that are closely associated with knowledge expression. 
This selection is made through gradient-based attributions, which effectively highlight the neurons that have a strong influence on the model's output.
After these critical neurons are identified, the method modifies the projection layer of the feed-forward network (denoted as $mlp^{(l)}_{proj}$) specifically at the rows corresponding to the selected neurons. 
This modification involves adding scaled embedding vectors to the current values, effectively adjusting the model's behavior in a targeted manner.
Specifically, they amplify knowledge neurons by doubling their activations. 
Similar to FT, all hyperparameters are adopted from default configurations(See code\footnote{https://github.com/EleutherAI/knowledge-neurons})

\paragraph{T-Patcher}
\label{sec:T-Patcher-Appendix}

The method proposed by \citet{huang2023transformerpatcher} offers a way to alter the behavior of transformer-based models with minimal changes.
Specifically, it adds and trains a small number of neurons in the last Feed-Forward Network (FFN) layer. 
This approach effectively provides a means for fine-tuning model behavior with less computational demand than comprehensive retraining.
It freezes all original parameters and adds one neuron (patch) to the last FFN layer for one mistake. 
And they train the patch to take effect only when encountering its corresponding mistake. 
For T5-XL implementation, all hyperparameters follow the same default settings as Bart-base\footnote{https://github.com/ZeroYuHuang/Transformer-Patcher}.

Furthermore, in the auto-regressive model (like GPT-J), the model may make multiple mistakes in one example. 
Therefore, for an example where the model makes n mistakes, we only consider errors generated by the model at the \textbf{output positions}. 
Following the settings of the original paper, we add up to 5 patches for one edit example. 
Formally, for an edit example $(x_e, y_e)$ in auto-regressive model,  the actual input is given by $\hat{x_e} = x_e + y_e$ and the patched model’s output is $p_e$, $l_e$ is defined as:

\begin{equation}
l_e = -\sum_{i=1}^{N} \hat{x_i} \log(p_i) \cdot \mathbf{1}_{(i \geq len(x_e))}
\end{equation}

\paragraph{ROME}
ROME, as proposed by ~\citet{meng2022locating}, conceptualizes the MLP module as a straightforward key-value store.
For instance, if the key represents a subject and the value encapsulates knowledge about that subject, then the MLP can reestablish the association by retrieving the value that corresponds to the key.
In order to add a new key-value pair, ROME applies a rank-one modification to the weights of the MLP, effectively ``writing in'' the new information directly. This method allows for more precise and direct modification of the model's knowledge.
We directly apply the code and MLP weight provided by the original paper \footnote{https://rome.baulab.info/} and keep the default setting for hyper-parameters.

\begin{table*}[ht]
    \centering
    \resizebox{1 \textwidth}{!}{
    \begin{tabular}{cll}
        \toprule
        Type     &  Edit Descriptor  & Portability Question  \\ 
        \midrule
        \multirow{3}{*}{\textbf{Subject Replace}} & In what living being can \emph{PRDM16} be found? &  In what living being can \emph{PR domain containing 16} be found? \\
        & When was \emph{Liu Song dynasty} abolished? & When was the end of \emph{the Former Song} dynasty? \\ 
        & \emph{Table tennis} was formulated in? & \emph{ping pang}, that originated in ?\\ \midrule
        \textbf{Reversed Relation} & What is Wenxiu's spouse's name?   & Who is the wife/husband of Wenxi Emperor?   \\ \midrule
        \textbf{One-hop Reason} & What company made Volvo B12M?   &  \makecell[l]{In which city is the headquarters of the company that \\ made the Volvo B12M?}  \\
        \bottomrule
    \end{tabular} 
    }
\caption{Example of portability dataset.}
\label{tab:port-dataset-example}
\end{table*}
\paragraph{MEMIT}
MEMIT~\cite{meng2023massediting} builds upon ROME to insert many memories by modifying the MLP weights of a range of critical layers. 
We test the ability of MEMIT using their code \footnote{https://memit.baulab.info/} and all hyperparameters follow the same default settings.
For GPT-J, we choose R = {3, 4, 5, 6, 7, 8}, and covariance statistics are collected using 100,000 samples of Wikitext.
For GPT-NEOX-20B, we select R = {6, 7, 8, 9, 10}, and covariance statistics are collected from over 50,000 samples of Wikitext.

\paragraph{IKE}
IKE~\cite{Zheng2023CanWE} defines three types of demonstration formatting templates including (i)copy, (ii)update, (iii)retain, which guide LMs to edit knowledge facts by in-context learning (ICL). As there are no parameter modifications, IKE is applicable to any existing LLMs. 

In alignment with the original paper, we choose k-NN examples from the training corpus(10000 size). The demonstrations are encoded by \texttt{all-MiniLM-L6-v2}. Following the default setting, we set k to 32(See code\footnote{https://github.com/PKUnlp-icler/IKE}).

\section{Dataset Details}
\label{app:dataset}
\subsection{Basic DataSet}
\textbf{ZsRE} ~\cite{levy-etal-2017-zero} is a Question Answering (QA) dataset using question rephrasings generated by back-translation as the equivalence neighborhood. 
\textbf{\cf}~\cite{meng2022locating} is a more challenging dataset that accounts for counterfacts that start with low scores in comparison to correct facts. 
It constructs out-of-scope data by substituting the subject entity for a proximate subject entity sharing a predicate. 
This alteration enables us to differentiate between superficial wording changes and more significant modifications that correspond to a meaningful shift in a fact.
We follow previous data split~\cite{de-cao-etal-2021-editing,meng2022locating,mitchell2022fast} to evaluate all the models on the test set.
For models requiring training, we utilize the training set. 
Following prior work~\cite{mitchell2022fast,Mitchell2022MemoryBasedME}, we use the Natural Questions (NQ; \citet{kwiatkowski-etal-2019-natural}) as out-of-scope data to evaluate locality.

\subsection{Dataset Construction for Portability Evaluation}
\subsubsection{One hop}
\begin{figure*}
    \centering
    \includegraphics[width=1\textwidth]{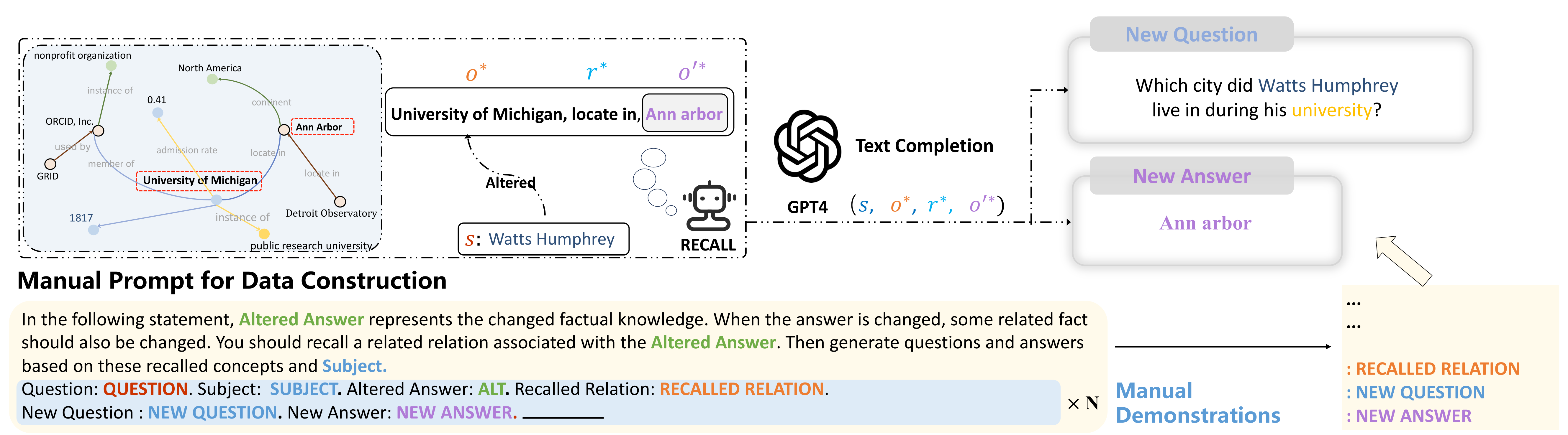}
    \caption{Dataset construction procedure to generate portability part (Q,A) with GPT4.}
    \label{fig:data_construction}
\end{figure*}

The construction can be seen in Figure~\ref{fig:data_construction}.
To ensure that the original model($f_\theta$) has seen the triple $(s, r, o)$ during the pre-training process, we employ link prediction to predict $o$ given $(s, r, ?)$. 
If the tail entity is present in the Top-10 logits, we consider the model to have prior knowledge of this triple. In other words, if the model has sufficient portability, it can correctly answer new questions based on the subject and the triplet.

We select data points to measure the performance of the model's portability. 
The symbolic representation of the portability dataset is as follows:

\begin{equation}
\nonumber
D_{port}=\{\text{GPT4}\left(s, r, o\right) \mid o \in \text{Top-}{10}
(f_{\theta}\left(s,r,?\right)) \}
\end{equation}

To guide GPT-4 in producing the desired question and answer, we employ a few-shot manual demonstration as a prompt (See Table \ref{tab:port-prompt}).
In addition, we intersect the data filtered by T5-XL and the data filtered by GPT-J to obtain the final portability dataset.
The GPTJ model achieves a link prediction score of ZSRE: \textbf{72.99} and \cf: \textbf{69.78}, while the T5 model achieves a link prediction score of ZSRE: \textbf{83.90} and \cf: \textbf{84.81}. It ensures that the models possess prior knowledge about this triple.

As a result, we select some data instances from the \textbf{ZsRE} and the \textbf{\cf} dataset. 
The description of the data is shown in Table \ref{tab:port-dataset-Statistics}.

\begin{table}[ht]
    \centering
    \resizebox{0.48\textwidth}{!}{
    \begin{tabular}{cccc}
        \toprule
             & \textbf{Subject Replace} & \textbf{Inverse Relation} & \textbf{One hop} \\ 
        \midrule
        \textbf{ZsRE} & 293 & 385  & 1,037 \\
        \textbf{\cf} & 213  & -  & 1,031  \\
        \bottomrule
    \end{tabular}
    }
\caption{Statistics of portability dataset.}
\label{tab:port-dataset-Statistics}
\end{table}
\begin{table*}[ht]
    \centering
    \resizebox{1 \textwidth}{!}{
    \begin{tabular}{cll}
        \toprule
        Type     &  Edit Descriptor  & Locality Question  \\ 
        \midrule
        \multirow{3}{*}{\textbf{Other Attribution}} & \emph{Grant Hill} is a professional \_ &  Which country does \emph{Grant Hill} represent in sport? (relation: \emph{country}) \\
        & The language of \emph{La Dispute} was \_ & What genre does \emph{La Dispute} belong to? (relation: \emph{genre}) \\ 
        & \emph{Gleb Kotelnikov} is a native speaker of \_ & What is the gender of \emph{Gleb Kotelnikov}? (relation: \emph{sex or gender})\\ \midrule
        \multirow{2}{*}{
        \textbf{Distract Neighbor}} & \emph{Windows 98} was a product of \_   &  \makecell[l]{\emph{Windows 98} was a product of IBM. Windows Media Center, developed by \_}  \\
        & The language of \emph{Goodfellas} is \_   &  \makecell[l]{The language of \emph{Goodfellas} is Tamil. The language of Titanic is \_}  \\
        \bottomrule
    \end{tabular} 
    }
\caption{Example of locality dataset.}
\label{tab:loc-dataset-example}
\end{table*}

\subsubsection{Subject Replace}
We replace the question’s subject with an alias or synonym to test generalization on other descriptions of the subject. We used two approaches to construct this dataset. 1. For subjects that could be found in Wikidata, we replaced the original subject with the alias from Wikidata (Field Name: \texttt{Also known as}). 2. For subjects that could not be found in Wikidata, we used GPT-4 to generate synonyms for the original subject.
This process ensures that the evaluation accurately reflects the model's capability to handle various subject representations, contributing to a more comprehensive understanding of its performance.
\subsubsection{Reversed Relation}
In an editing instance, the attributes of the target entity can also change. For instance, in the edited instance: "Who is the father of Nebaioth? Ishmael $\rightarrow$ Babur." When answering the question "Who is the son of Babur?" it should be answered based on the new fact after editing, which is \textbf{Nebaioth}.
Certain types of relations may not be as effective for evaluation. Let's consider a hypothetical scenario where we change the location of the Eiffel Tower to Rome - proposing a valid reversed question in such a context would be challenging.
Consequently, we carefully handpicked all relations in the ZsRE dataset that could be reversed, such as one-to-one relations, and selected related questions through keywords (such as \texttt{spouse, wife, mother, father, brother, sister}) screening.  This allows us to maintain the integrity and relevance of the evaluation process, thus ensuring more reliable results.
To guide GPT-4 in producing the desired reversed question, we employ a few-shot manual demonstration as a prompt (See Table \ref{tab:inverse-prompt}).

\subsection{Dataset Construction for Locality Evaluation}
\label{sec:locality_cons}
\subsubsection{Other Attribution}
Other attributes of the subject updated should remain the same before editing. For example, if we edit basketball player \emph{Grant Hill} as a soccer player, it does not affect his nationality. Therefore, for unrelated attributes like \emph{country}, the output should remain consistent with the pre-editing version. We modified the \textbf{\cf} dataset by using the Wikidata API to traverse all relationships of a subject and randomly select an unrelated relationship and tail entity as a data sample. We provide $(s, r_{other})$ to GPT-4 to generate a question, and the answer to this question corresponds to the respective tail entity. As a result, we modify \textbf{804} data instances from the \textbf{\cf} dataset.

\subsubsection{Distract Neighbor}
Following \citet{jason2023detecting}, we modify the neighborhood prompt in \textbf{\cf} dataset by prepending the model edit. For example(See Table \ref{tab:loc-dataset-example}), if the original prompt is "Windows 98 was a product of \_" the modified prompt would be "Windows 98 was a product of IBM. Windows Media Center, developed by \_". It measures whether the model editing technique has resulted in significant side effects on the model itself due to over-editing. As a result, we select \textbf{804} data instances from the \textbf{\cf} dataset.

\subsubsection{Other Task}
We select commonsense tasks here to assess the post-edited model's performance on other tasks.
Given a question \emph{q}, multiple-choice commonsense reasoning aims to select the correct answer $a_t \in \mathcal{A}$ provided with an optional context \emph{c}. \textbf{Physical Interaction QA} (PIQA (\cite{bisk2020piqa}) is a 2-way multiple-choice QA task testing physics reasoning about objects. We evaluate the post-edit model on the PIQA dataset to reflect the impact of different model editing techniques on the performance of other downstream tasks. Specifically, For each model editing technique, we \textbf{sequentially} edit GPT-J with 100 samples in the \textbf{\cf} dataset. Afterward, we test the performance of the continuously post-edit models on PIQA, using accuracy as the selected metric, which is defined as :

\begin{equation}
\emph{acc} = \sum_{k=1}^{N} Q(c_k, q_k, a_{k_p}) / N
\end{equation}

where $a_{k_p}$ is the option with the least perplexity of the post-edit model, $Q(c_k, q_k, a_{k_p})$ is 1 if $a_{k_p} = a_{k_t}$ and 0 otherwise.

\begin{table*}[ht]
\centering
\begin{tabular}{l}
\textbf{Prompt} \\ 
\midrule
\begin{tabular}[c]{@{}l@{}} 
In the following statement, `Altered Answer` represents the changed factual knowledge. \\
When the answer is changed,  some related facts should also be changed. You should \\

recall a related relation associated with the `Altered Answer`. Then generate questions \\

and answers based on these recalled concepts and `Subject`. \\ \\ \end{tabular} \\

\begin{tabular}[c]{@{}l@{}} 
\midrule
Question: What university did Watts Humphrey attend? \\
Subject: Watts Humphrey \\
Altered Answer: University of Michigan \\
\textbf{Recalled Relation: (University of Michigan, locate in, Ann Arbor)} \\
New Question: Which city did Watts Humphrey live in during his \\ undergraduate studies? \\
New Answer: Ann Arbor in Michigan State \\ \\
\midrule
Question: Windows 10, developed by \\
Subject: Windows 10 \\
Altered Answer: Google \\
\textbf{Recalled Relation: (Sundar Pichai, ceo of, Google)} \\
New Question: Who is the CEO of the company that develops the Windows 10 operating system? \\ 
New Answer: Sundar Pichai \\ \\
\midrule
Question: In Kotka, the language spoken is? \\
Subject: Kotka \\
Altered Answer: French \\
\textbf{Recalled Relation: (French, evolve from, Romance)} \\
New Question: What language did Kotka's official language evolve from? \\
New Answer: Romance \\ \\
\midrule
Question: Armand Trousseau’s area of work is? \\
Subject: Armand Trousseau \\
Altered Answer: jazz \\
\textbf{Recalled Relation: (Miles Davis, genres, jazz)} \\
New Question: Armand Trousseau formed a band during college, they are all fans of? \\
New Answer: Miles Davis
\end{tabular}
\end{tabular}
\caption{Prompt for dataset construction on zsRE \& \textsc{CounterFact} portability dataset. Demonstration examples are manually constructed. For each data instance, we provide Question, Subject, and Altered Answer to generate portability data.}
\label{tab:port-prompt}
\end{table*}

\begin{table*}[ht]
\centering
\begin{tabular}{r l}
\textbf{Task} & \textbf{Prompt} \\ \hline
\begin{tabular}[c]{@{}l@{}} \textbf{ZSRE} \\ \\ \end{tabular} & \begin{tabular}[c]{@{}l@{}} Please generate the Inverse Question(For example, A and B are in a father-son relationship. \\ In the original question, it says "who is the father of B? Answer is A". You should ask who \\ is the son/daughter of A, so answer is B.) here are some examples: \\ \\ \end{tabular} \\
 & \begin{tabular}[c]{@{}l@{}}
Q: Who is Claire Clairmont's sister? A: Marian Clairmont \\
Inverse Question: Who is Marian Clairmont's sister? \\ \\
Q: What was the name of the father of Jane Seymour? A: Richard Seymour \\
Inverse Question: Who is the son/daughter of Richard Seymour? \\ \\ 
Q: What is Elizabeth Grey, Countess of Kent's spouse? A: Henry Grey, 1st Duke of Suffolk \\
Inverse Question: Who was Henry Grey, 1st Duke of Suffolk married to? \\ \\
Q: Who is listed as Leonor, Princess of Asturias's father? A: Leonor III of Spain \\
Inverse Question: Who is the son/daughter of Leonor III of Spain?\end{tabular}
\end{tabular}
\caption{Prompt for inversed relation dataset construction on zsRE, we provide Question and Answer to generate an inversed question.}
\label{tab:inverse-prompt}
\end{table*}

\end{document}